\documentclass[sigconf]{acmart}

\usepackage{amsmath,amssymb,amsfonts}
\usepackage{algorithmic}
\usepackage{graphicx}
\usepackage{textcomp}
\usepackage{xcolor}
\usepackage{times}
\usepackage{array,booktabs,caption,ragged2e}
\usepackage{amsfonts}
\usepackage[english]{babel}
\usepackage{multirow}
\usepackage{url}
\usepackage{subfigure}
\usepackage{amsmath,amsfonts,bm}

\def\eqref#1{equation~\ref{#1}}
\def\1{\bm{1}}

\def\vb{{\bm{b}}}
\def\vc{{\bm{c}}}

\def\vf{{\bm{f}}}

\def\vi{{\bm{i}}}

\def\vm{{\bm{m}}}

\def\vp{{\bm{p}}}

\def\vr{{\bm{r}}}

\def\vt{{\bm{t}}}

\def\mA{{\bm{A}}}

\def\mD{{\bm{D}}}

\def\mF{{\bm{F}}}

\def\mI{{\bm{I}}}

\def\mL{{\bm{L}}}
\def\mM{{\bm{M}}}

\def\mR{{\bm{R}}}

\def\mT{{\bm{T}}}

\def\mW{{\bm{W}}}

\DeclareMathAlphabet{\mathsfit}{\encodingdefault}{\sfdefault}{m}{sl}
\SetMathAlphabet{\mathsfit}{bold}{\encodingdefault}{\sfdefault}{bx}{n}
\newcommand{\tens}[1]{\bm{\mathsfit{#1}}}

\def\tC{{\tens{C}}}

\def\tG{{\tens{G}}}
\def\tH{{\tens{H}}}

\def\tM{{\tens{M}}}

\def\tR{{\tens{R}}}
\def\tS{{\tens{S}}}

\def\tU{{\tens{U}}}

\def\sH{{\mathbb{H}}}

\def\sP{{\mathbb{P}}}

\def\sR{{\mathbb{R}}}

\def\sV{{\mathbb{V}}}

\hypersetup{draft}
\setcopyright{rightsretained}
\begin{document}
%\title[OD Forecasting]{Stochastic Origin-Destination Matrix Forecasting Using Dual-Stage Graph Convolutional, Recurrent Neural Networks}
\title[Recurrent Multi-Graph Neural Networks for Travel Cost Prediction]{Recurrent Multi-Graph Neural Networks for Travel Cost Prediction}
%\acmConference{ACM Conference}{NY}{USA}
%\titlenote{Produces the permission block, and
%  copyright information}
%\subtitle{Extended Abstract}
%\subtitlenote{The full version of the author's guide is available as
%  \texttt{acmart.pdf} document}
%\author{Paper ID 1335}
\author[Hu, Guo, Yang, Jensen, Chen]{Jilin Hu,~~~Chenjuan Guo,~~~Bin Yang,~~~Christian S. Jensen,~~~Lu Chen \\
Department of Computer Science, Aalborg University, Denmark\\
\texttt{\{hujilin, cguo, byang, csj, luchen\}@cs.aau.dk} \\}

\begin{abstract}

%Given a partitioning of a road network into regions, an origin-destination (OD) matrix records the cost of travel between any pair of regions, example costs being travel speed or greenhouse gas emission. OD matrices have a broad range of uses in transportation and logistics. We consider an increasingly pertinent setting where a set of vehicle trajectories is used for instantiating historical OD matrices. As a cost such as travel speed varies over time, e.g., due to varying congestion, matrices are created for different time intervals during a day, e.g., one matrix for every 15 minutes. A cost is modeled as a distribution because different vehicles, as captured by the set of trajectories, may travel at different speeds during the same time interval, e.g., due to different driving styles or different waiting times at intersections. The resulting historical OD matrices are likely to be sparse. We address the problem of forecasting complete near future OD matrices from such sparse historical OD matrices. To solve the problem, we propose a generic learning framework that employs matrix factorization and graph convolutional neural networks to contend with data sparseness and that captures temporal dynamics via recurrent neural networks. Empirical studies using two taxi trajectory data sets offer detailed insight into the properties of the framework and indicate that it is effective.

Origin-destination (OD) matrices are often used in urban planning, where a city is partitioned into regions and an element $(i$, $j)$ in an OD matrix records the cost (e.g., travel time, fuel consumption, or travel speed) from region $i$ to region $j$. In this paper, we partition a day into multiple intervals, e.g., 96 15-min intervals and each interval is associated with an OD matrix which represents the costs in the interval; and we consider sparse and stochastic OD matrices, where the elements represent stochastic but not deterministic costs and some elements are missing due to lack of data between two regions.

We solve the sparse, stochastic OD matrix forecasting problem. Given a sequence of historical OD matrices that are sparse, we aim at predicting future OD matrices with no empty elements. We propose a generic learning framework to solve the problem by dealing with sparse matrices via matrix factorization and two graph convolutional neural networks and capturing temporal dynamics via recurrent neural network. Empirical studies using two taxi datasets from different countries verify the effectiveness of the proposed framework.

%Urban planners are seeking for a smart traffic monitor system which provides a high-resolution traffic speed within a region of interests~(RoIs). In particularly, travel speeds within RoIs is modeled as a time-dependent distribution: given one pair of origin and destination subregion, there may exist several trips at the same time period, of which has its own unique travel speed such that all the travel speed of those trips can be collected to construct a travel speed distribution. Such origin destination~(OD) stochastic speed can be collected from floating car dataset. However, it is still an almost impossible task to collect such a big dataset that covers all the OD pairs at all times which contradicts the desire of transport planners who are eager to have a high-resolution stochastic speed in an RoIs currently and in the future.
%
%We solve the problem of OD stochastic speed forecasting. To better solve the data sparseness and spatial correlation problem, we formalize this problem as low-rank matrix approximation and geometric structure among the subregions in RoIs. We propose a generic learning framework to forecast the OD stochastic speed with matrix fatorizations and capturing the temporal dynamics via recurrent neural network. Next, we consider the spatial correlations for both origin and destination dimension for better factorizations. Empirical studies using two taxi datasets from two countries verify the effectiveness of the proposed framework.
\end{abstract}

%
% The code below should be generated by the tool at
% http://dl.acm.org/ccs.cfm
% Please copy and paste the code instead of the example below.
%
%\begin{CCSXML}
%<ccs2012>
% <concept>
%  <concept_id>10010520.10010553.10010562</concept_id>
%  <concept_desc>Computer systems organization~Embedded systems</concept_desc>
%  <concept_significance>500</concept_significance>
% </concept>
% <concept>
%  <concept_id>10010520.10010575.10010755</concept_id>
%  <concept_desc>Computer systems organization~Redundancy</concept_desc>
%  <concept_significance>300</concept_significance>
% </concept>
% <concept>
%  <concept_id>10010520.10010553.10010554</concept_id>
%  <concept_desc>Computer systems organization~Robotics</concept_desc>
%  <concept_significance>100</concept_significance>
% </concept>
% <concept>
%  <concept_id>10003033.10003083.10003095</concept_id>
%  <concept_desc>Networks~Network reliability</concept_desc>
%  <concept_significance>100</concept_significance>
% </concept>
%</ccs2012>
%\end{CCSXML}
%
%\ccsdesc[500]{Computer systems organization~Embedded systems}
%\ccsdesc[300]{Computer systems organization~Redundancy}
%\ccsdesc{Computer systems organization~Robotics}
%\ccsdesc[100]{Networks~Network reliability}

%\keywords{ACM proceedings, \LaTeX, text tagging}

\maketitle

\section{Introduction}

Origin-destination (OD) matrices~\cite{ekowicaksono2016estimating,DBLP:journals/vldb/HuYGJ18} are applied widely in location based services and online map services (e.g., transportation-as-a-service), where OD matrices are used for the scheduling of trips, for computing payments for completed trips, and for estimating arrival times. For example, Google Maps\footnote{\url{https://tinyurl.com/7vmtk4y}} and ESRI ArcGIS Online\footnote{\url{https://tinyurl.com/ydhq765h}} offer OD matrix services to help developers to develop location based applications. Further, increased urbanization contributes to making it increasingly relevant to capture and study city-wide traffic conditions. OD matrices may also be applied for this purpose.

\begin{figure}[t]
	\centering
	\subfigure[Grid-based Partition]{
		\begin{minipage}[b]{0.45\linewidth}
			\includegraphics[width=1\linewidth]{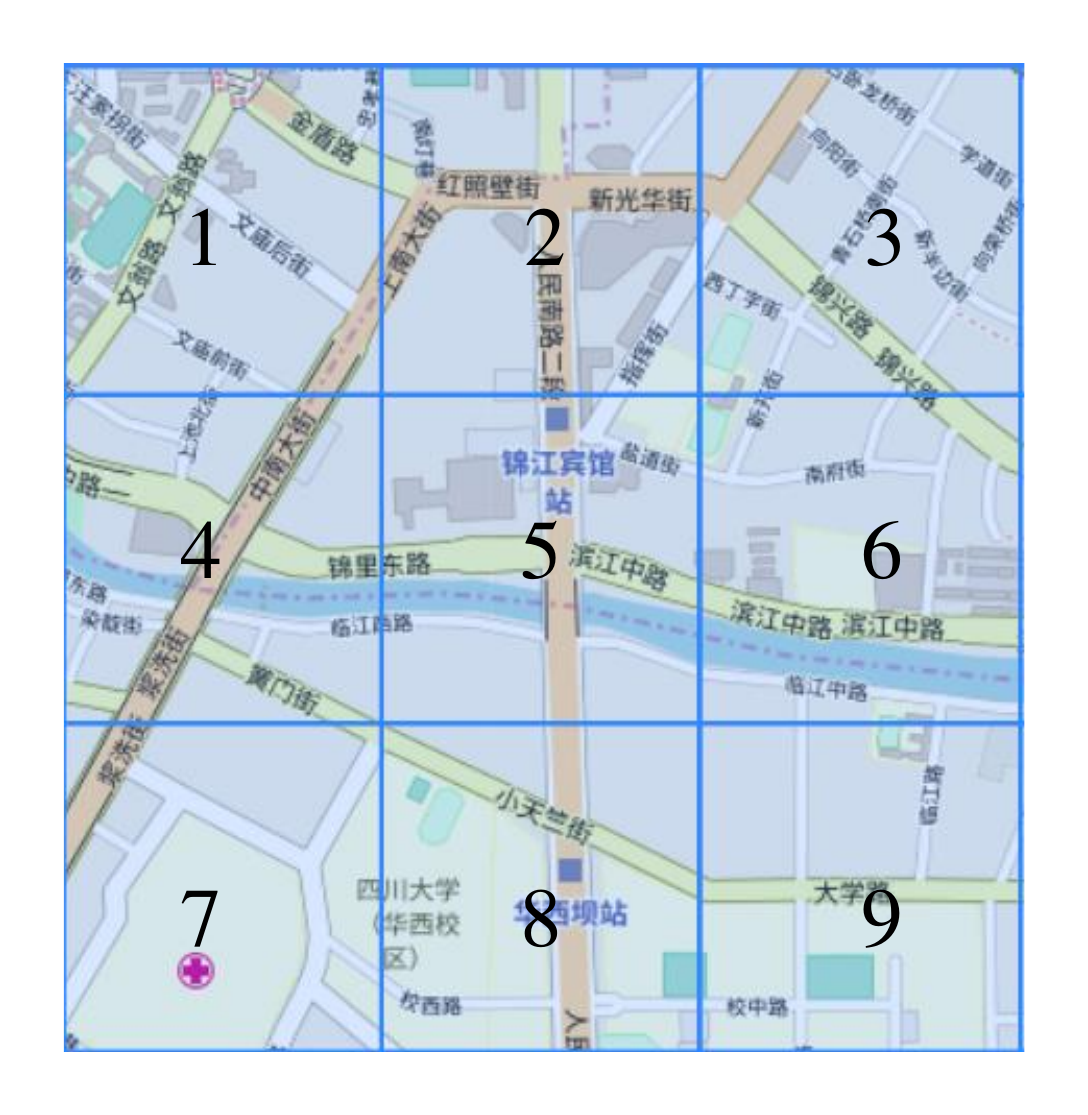}
		\end{minipage}
		\label{fig:grid_rep}
	}
	\subfigure[Road-based Partition]{
		\begin{minipage}[b]{0.45\linewidth}
			\includegraphics[width=1\linewidth]{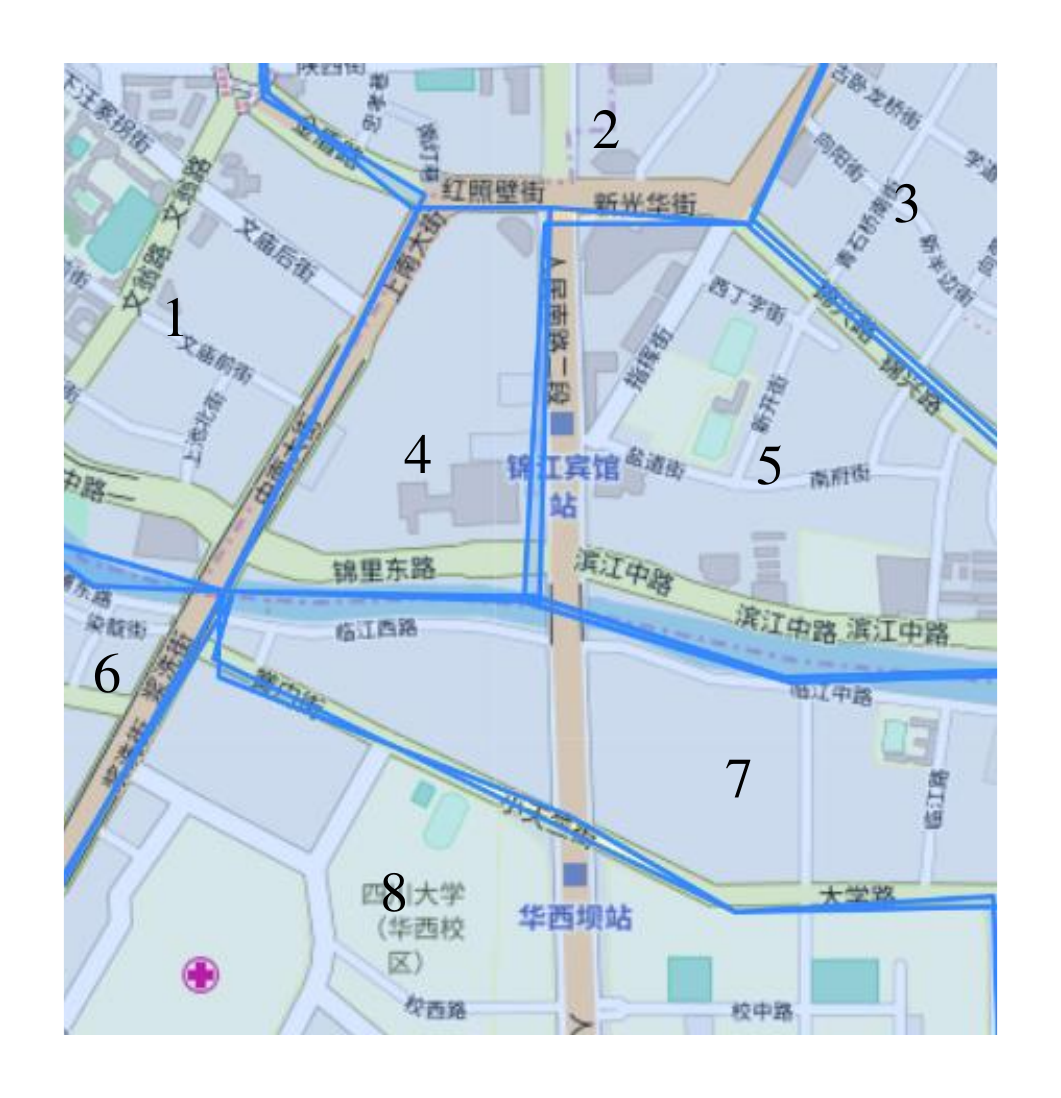}
		\end{minipage}
		\label{fig:poly_rep}
	}
	\vspace{-10pt}
	\caption{Partition a City into Regions}
	\label{fig:intro_fig}
	\vspace{-10pt}
\end{figure}

To use OD-matrices, a city is partitioned into \textit{regions}, and a day is partitioned into \textit{intervals}. 
Each interval is assigned its own an OD-matrix, and an element $(i$, $j)$ in matrix described the attribute (e.g., travel speed, fuel consumption~\cite{DBLP:journals/geoinformatica/Guo0AJT15,DBLP:conf/gis/GuoM0JK12}, or travel demand) of travel from region $i$ to region $j$ during the interval that the matrix represents.  
Different approaches can be applied to partition a road network~\cite{DBLP:journals/tc/Ding0C016,DBLP:conf/dasfaa/YangMQZ09}, e.g., using a uniform grid or using major roads, as exemplified in Figure~\ref{fig:intro_fig}. In this paper, we focus on speed matrices. However, the proposed techniques can be applied on other travel attributes or costs, such as travel time, fuel consumption, and travel demand. 

As part of the increasing digitization of transportation, increasiingly vast volumes of vehicle trajectory, trajectory data are becoming available~\cite{DBLP:journals/sigmod/GuoJ014,DBLP:journals/tits/Ding0GL15}. We aim to exploit such data for composing OD matrices.
Specifically, an element $(i$, $j)$ of a speed matrix for a given time interval can be instantiated from the speeds observed in trajectories that went from region $i$ to region $j$ during the relevant time interval. 

We consider \textit{stochastic OD matrices} where the elements represent uncertain costs by meaning of cost distributions rather than deterministic, signle-valued costs. The use of distribution models reality better and enables more reliable decision-making. 
For example, element $(i$, $j)$ has a speed histogram $\{([10, 20), 0.5)$, $([20, 40), 0.3)$, $([40, 60], 0.2)\}$, meaning that the probability of traveling speed from region $i$ to region $j$ at 0-20~km/h is 0.5, at $[20, 30)$ is 0.3, and at $[30, 40)$ is 0.2, respectively. If a passenger needs to go from his home in region $i$ to catch a flight in an airport in region $j$, and the shortest path from his home to the airport is 20~km, then we are able to derive a travel time (minutes) distribution: $\{[30, 40], 0.5)$, $(40, 60], 0.3)$, $(60, 120], 0.2)\}$. Therefore, the passenger needs to reserve at least 120 minutes for not being late. However, when only using average speed to derive an average travel time of 54 minutes, it makes the passenger runs into a risk of missing the flight.  

%For example, based on the speed distribution between region $i$ and region $j$, we are able to derive the travel time distributions, for example $\{(100, 0.6)$, $(120, 0.4)\}$, meaning that it may take 100 or 120 minutes to travel from region $i$ to region $j$ with probabilities 0.6 and 0.4, respectively. If a person needs to go to region $j$ to catch a flight, to make sure on time arrivial, at least 120 mins need to be reserved. Only considering average travel time, i.e., 108 mins, runs into a risk of missing the flight. 
%For example, based on the speed distribution between region $i$ and region $j$, we are able to derive the travel time distributions, for example $\{(100, 0.6)$, $(120, 0.4)\}$, meaning that it may take 100 or 120 minutes to travel from region $i$ to region $j$ with probabilities 0.6 and 0.4, respectively. If a person needs to go to region $j$ to catch a flight, to make sure on time arrivial, at least 120 mins need to be reserved. Only considering average travel time, i.e., 108 mins, runs into a risk of missing the flight. 

\begin{figure}[htb]
	\begin{center}
		\includegraphics[width=0.95\linewidth]{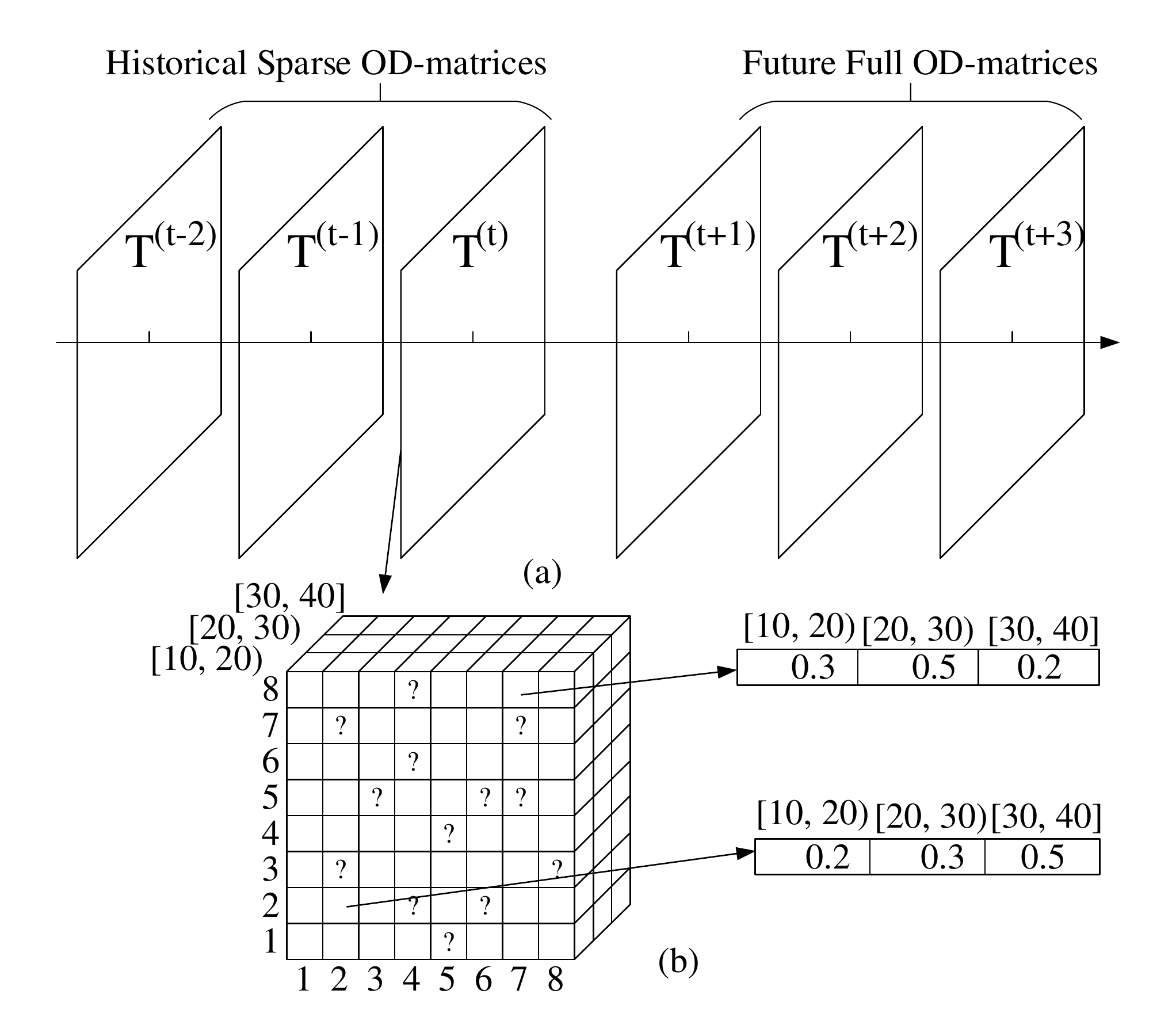}
	\end{center}
	\caption{Stochastic Origin-Destination Matrix Forecasting}
	\label{fig:mot}
	\vspace{-10pt}
\end{figure}

We address the problem of \textit{stochastic origin-destination matrix forecasting}---based on historical stochastic OD-matrices, we predict future OD-matrices. Figure~\ref{fig:mot} shows a specific example: given stochastic OD-matrices for 3 historical intervals $T^{(t-2)}$, $T^{(t-1)}$, and $T^{(t)}$, we aim at predicting the stochastic OD-matrices for the 3 future intervals $T^{(t+1)}$, $T^{(t+2)}$, and $T^{(t+3)}$. 
%
%In this paper, we propose a origin-destination~(OD) stochastic speed forecast problem. Firstly, we construct the OD stochastic speed as a 3-D tensor that each  element is a corresponding stochastic weight. Continue with the geographical representation in Figure~\ref{fig:poly_rep}, we construct a correponding running example to demonstrate the problem more obviously. 

Here, a stochastic OD-matrix is represented as a 3-dimensional tensor, where the first dimension represents source regions, the second dimension represents destination regions, and the third dimension represents cost ranges. 
For example, Figure~\ref{fig:mot}(b) shows the stochastic OD-matrix for interval $T^{(t)}$, which is represented as a $\mathcal{R}^{8\times8\times3}$ tensor with 8 source regions, 8 destination regions, and 3 speed (km/h) ranges [10, 20), [20-30), and [30-40]. 
Element (8, 8) in the OD-matrix is a vector (0.3, 0.5, 0.2), meaning that, when traveling within region 8, the travel speed histogram is $\{([10, 20), 0.3)$, $([20, 30), 0.5)$, $([30, 40], 0.2)\}$.

Solving the stochastic OD-matrix forecasting problem is non-trivial as it is necessary to contend with two difficult challenges. \\
(1)~\textbf{Data Sparseness.} To instantiate a stochastic OD-matrix in an interval using trajectories, we need to have sufficient trajectories for each region pair during the interval. However, even massive trajectory data sets are often spatially and temporally skewed~\cite{simlemethod_zhenhui, ST_NN, Yuzheng_TTE, Wang_zheng_lreta,DBLP:conf/icde/Guo0HJ18}, making it almost impossible to cover all region pairs for all intervals. 

For example, the New York City taxi data set\footnote{\url{http://www.nyc.gov/html/tlc/html/technology/raw_data.shtml}} we use in experiments has more than 29 million trips from November and December 2013. Yet, this massive trip set only covers 65\% of all ``taxizone'' pairs in Manhattan, the most densely traversed region in New York City.  If we further split the data set according to the temporal dimension, e.g., into 15-min intervals, the spareness problem becomes even more severe.

The data sparseness in turn results in sparse historical stochastic OD-matrices, where some elements are empty (e.g., those elements with ``?'' in Figure~\ref{fig:mot}(b)). Yet, decision making requires full OD-matrices. The challenge is how to use sparse historical OD-matrices to predict full future OD-matrices.

%Firstly, although more and more vehicles are equiped with GPS devices nowadays, it is still difficult to collect those data due to the privacy problem. Secondly, even though we can collect millions of observations from the running taxis, the observations cannot cover all the region pairs in the city. For example, we have 29.3~million trips in November and December 2013, it still cannot cover all the region pairs where the regions are divided by the segmentation shapefile named Taxizone--More than 35\% regions pairs in the borough of Manhattan do not covered by any trips. However, Manhattan is the borough with the densest trips in NYC-taxi dataset, the situation in other boroughs will be even worse. Furthermore, if we split those observations by temporal, e.g., splitting into 15 minutes as a time interval and we accumulate all the data during this time interval, the data spareness problem will be much more severe.

\noindent
(2)~\textbf{Spatio-temporal Correlations.} Traffic is often spatio-temporally correlated---if a region is congested during a time interval, its neighboring regions are also likely to be congested in subsequent intervals. 
Thus, to predict accurate OD-matrices, we need to account for such spatio-temporal correlations.  
However, the OD-matrices themselves do not necessarily capture spatial proximity. 
No matter which partition method is used, we cannot always guarantee that two gegraphically adjacent regions are represented by adjacent rows and columns in the matrix. For example, in Figure~\ref{fig:grid_rep}, regions 1 and 4 are geographically adjacent, but they are not adjacent in the OD matrices; in Figure~\ref{fig:poly_rep}, regions 4 and 7 are adjacent but they are again not adjacent in the OD matrices. 
%
%the $i$-th region and the $(i+1)$-th region may not be spatially close to each other but be further away. 
%
This calls for a separate mechanism that is able to take into account the geographical proximity of regions.

We propose a data-driven, end-to-end deep learning framework to forecast stochastic OD matrices that aims to effectively address the challenges caused by data sparseness and spatio-temporal correlations. 
First, to address the data sparseness challenge, we factorize a sparse OD matrix into two small dense matrices with latent features of the source regions and the destination regions, respectively. 
Second, we model the spatial relationships among source regions and among destination regions using two graphs, respectively.  Then, we employ two graph covolutional, recurrent neural networks (GR) on the two dense matrices to capture the spatio-temporal correlations. 
Finally, the two GRs predict two dense, small matrices. We apply the multiplication to the two dense, small matrices to obtain a full predicted OD-matrix.

%We first model the spatial connection among the regions via graph. Then, we feed the data into the model to learn the latent representation for origin and destination regions, respectivelly. After it, the learned latent representation is propagated via graph convolutional recurrent neural network~(GR) to capture both spatial correlation and temporal dynamics. Finally, we recover the full forecasting tensor from the latent representation of origin and destination with matrix factorization. The proposed framework is generic in the sense that it can be further extended to fulfill the task of tensor completion or forecasting with graph spatial connection on both x and y axis. The results in our paper also show its performance are superior to the traditional machine learning methods, e.g., support vector regression~(SVR), and general deep learning based method, e.g., fully connected recurrent neural network~(FCR). 

To the best of our knowledge, this is the first study of stochastic OD matrix forecasting that contends with data sparseness and spatio-temporal correlations. The study makes four contributions. First, it formalizes the stochastic OD matrix forecasting problem. Second, it proposes a generic framework to solve the problem based on matrix factorization and recurrent neural networks.   
Third, it extends the framework by embedding spatial correlations using two graph convolutional neural networks. Fourth, it encompasses an extensive experiments using two real-world taxi datasets that offers insight into the effectiveness of the framework.

The remainder of the paper is organized as follows. Section 2 covers related works. Section 3 defines the setting and formalizes the problem. Section 4 introduces a basic framework and Section 5 presents an advanced framework. %h is extended to account for spatio-temporal correlations in Section 5.% proposes an advanced model that considers spatial dependency. 
Section 6 reports experiments and Section 7 concludes.

\section{Related Work}
\subsection{Travel Cost Forecasting}
We consider three types of travel cost forecasting methods in turn: segment-based methods~\cite{rice2001simple,DBLP:conf/kdd/YuanZXS11,DBLP:journals/pvldb/0002GJ13,DBLP:journals/tits/WuHL04,DBLP:conf/icde/YangGJKS14,DBLP:journals/vldb/0002GMJ15,DBLP:journals/geoinformatica/HuYJM17}, path-based methods~\cite{DBLP:journals/vldb/YangDGJH18,DBLP:journals/pvldb/DaiYGJH16,Yuzheng_TTE,Wang_zheng_lreta, AAAI_18_WDONG,zhang2018deeptravel}, and OD-based methods~\cite{simlemethod_zhenhui, li2018multi, ST_NN}. 

Segment-based methods focus on predicting the travel costs of individual road segments. 
%Fixed sensors, e.g., Bluetooth or loop detectors, are deployed on road segments to monitor the traffic of the road segments, e.g., recording the travel costs of the road segments. Floating car data, e.g., GPS data, can also be employed. 
%
%
%
%Some studies model 
For example, by modeling the travel costs of a road segment as a time series, techiniques such as time-varying linear regression~\cite{rice2001simple}, Markov models~\cite{DBLP:conf/kdd/YuanZXS11,DBLP:journals/pvldb/0002GJ13}, and support vector regression~\cite{DBLP:journals/tits/WuHL04} can be applied to predict future travel costs. Most such models consider time series from different edges independently. As an exception, the spatio-temporal Hidden Markov model~\cite{DBLP:journals/pvldb/0002GJ13} takes into account the correlations among the costs of different edges. % except the
%
% To cope with it, a spatio-temporal Hidden Markov model~\cite{DBLP:journals/pvldb/0002GJ13} was proposed to predict travel costs (both travel times and greenhouse gas emissions) of multiple road segments while taking into account the spatio-temporal correlations among the travel costs from the road segments. 
Some other studies focus on estimating high-resolution travel costs, % of individual road segments, 
such as uncertain costs~\cite{DBLP:conf/icde/YangGJKS14,DBLP:journals/geoinformatica/HuYJM17} and personalized costs~\cite{DBLP:journals/vldb/0002GMJ15,DBLP:conf/icde/DaiYGD15}. The data sparseness problem has also been studied---methods exist to estimate travel costs for segments without traffic data~\cite{icde2019,DBLP:journals/tkde/YangKJ14}.  

Path-based methods focus on predicting the travel costs of paths. A naive approach is to predict the costs of the edges in a path and then aggregate the costs. However, this approach is inaccurate since it ignores the dependencies among the costs of different edges in paths~\cite{DBLP:journals/pvldb/DaiYGJH16, Yuzheng_TTE}. 
Other methods~\cite{DBLP:journals/vldb/YangDGJH18,DBLP:journals/pvldb/DaiYGJH16,Yuzheng_TTE} use sub-trajectories to capture such dependencies and thus to provide more accurate travel costs for paths. In particular, the PAth-CEntric (PACE) model is proposed to utilize sub-trajectories that may overlap to achieve the optimal accuracy~\cite{DBLP:journals/vldb/YangDGJH18,DBLP:journals/pvldb/DaiYGJH16}, whereas the other study only considers non-overlapping sub-trajectories~\cite{Yuzheng_TTE}. 
A few studies propose variations of deep neural networks~\cite{Wang_zheng_lreta, AAAI_18_WDONG,zhang2018deeptravel} to enable accurate travel-time prediction for paths. 

Finally, OD-based methods aim at predicting the travel cost for given OD pairs. 
Our proposal falls into this category. 
A simple and efficient baseline~\cite{simlemethod_zhenhui} is to compute a weighted average over all historical trajectories that represent travel from the origin to the destination in an OD pair. However, it does not address data sparseness, which means that if no data is available for a given OD pair, it cannot provided a prediction. In contrast, our proposal is able to predict full OD-matrices without empty elements based on historical, sparse OD-matrices.  
A recent study~\cite{li2018multi} utilizes deep learning and multi-task learning to predict OD travel time while taking into account considers the road network topology and the paths used in the historical trajectories. However, path information may not always be available. An example is the New York taxi data set that we use in the experiments. This reduces the applicability of the model. In contrast, our proposal does not require path information. 
Further, existing proposals support only deterministic costs, while our proposal also supports stochastic costs.   

\subsection{Graph Convolutional Neural Network}
Convolutional Neural Networks (CNNs) have been used successfully in the contexts of images~\cite{convolutional_2012}, videos~\cite{le_learning_2011}, speech~\cite{hinton_deep_2012}, 
%taxi supply-demand~\cite{DBLP:conf/icde/WangCLY17}, 
time series~\cite{DBLP:conf/cikm/CirsteaMMG018,DBLP:conf/mdm/Kieu0J18}, and trajectories~\cite{DBLP:conf/cikm/Kieu0GJ18}, where the underlying data is represented as a matrix~\cite{defferrard_convolutional_2016,bruna_spectral_2013}. For example, when representing an image as a matrix, nearby elements, e.g., pixels, share local features, e.g., represent parts of the same object. 
In constrast, in our setting, an OD-matrix may not satisfy the assumption that helps make CNNs work---two adjacent rows in an OD matrix may
represent two geographically distant regions and may not share any features; and two separated rows in an OD matrix may represent geographically close regions that share many features. 
%
%However, CNNs are insufficient for capturing local features among adjacent vertices in a graph~\cite{defferrard_convolutional_2016,bruna_spectral_2013}.
%[To Bin: I replace the following sentence with the above sentence.] these approaches only consider the local features along the spatial coordinates of the grids, and they do not consider the geometry of the local features~\cite{bruna_spectral_2013}.

Graph convolutional neural networks (GCNNs)~\cite{defferrard_convolutional_2016,bruna_spectral_2013} aim to address this challenge. 
In particular, the geographical relationships among regions can be modeled as a graph, and GCNNs then take into account the graph while learning.  
%
%
%as a special type of neural networks for graphs by introducing a spectral construction.
%
%\cite{defferrard_convolutional_2016} proposes a Chebyshev expansion to fasten the spectral convolution and a coarsening method to speed up the pooling operation [I don't feel this paper is relevant to our paper?? delete?].
%
One study~\cite{kipf_semi-supervised_2016} applies GCNNs to solve semi-supervised classification in the setting of citation networks and knowledge graphs. One study continues to study semi-supervised classification via dual graph convolutional networks~\cite{DBLP:conf/www/Zhuang018}.  %However, it has only one dimension with spatial correlations which can be applied with GCNNs and it does not have the characteristic of temporal dynamics. 
Another study~\cite{li_graph_2017} constructs GCNNs together with a Recurrent Neural Network (RNN) to forecast traffic and one recent study~\cite{icde2019} utilizes GCNNs to fill in travel time for edges without traffic data. 
All the above studies consider a setting where only one dimension needs to be modeled as a graph. In contrast, in our study, both dimensions, i.e., the source region dimension and the destination region dimension, need to be modeled as two graphs.  
%Both studies only invovle one dimension with spatial correlations which can be applied with GCNNs and it does not have the characteristic of temporal dynamics. 
%
An additional, recent study focuses on so-called geomatrix completion which considers a similar setting where two dimensions need to be modelded as two graphs. It uses multi-graph neural networks~\cite{DBLP:conf/nips/MontiBB17} with RNNs. However, the RNNs in this study are utilized to perform iterations to approximate the geomatrix completion, not to capture temporal dynamics as in our study. 
To the best of our knowledge, our study is the first that constructs a learning framework involving dual-graph convolution and employing RNNs to forecast the future. %stochastic speed ODs.

\section{Preliminaries}

\subsection{OD Stochastic Speed Tensor}
\label{sect:pre_od}
%A {\bf vertex } $v$ in a road network is defined as an intersection in a road network and all the intersections consist of a vertex set $\sV$. 
%In addition, a road network can be partitioned into $N$ regions geographically such that we can have $N$ subsets of vertex, i.e., $\sV = \{\sV_1, \cdots, \sV_N\}$. 

%A {\bf trip} $\vp$ is defined as a tuple $\displaystyle \vp=(o, d, t, l, \tau)$, where $o, d \in \sV$ denote origin and destination vertices, $t$ is a departure time, $\displaystyle l$ represents the trip distance, and $\tau$ is the travel time of the trip. Given $\vp.l$ and $\vp.\tau$, we can derive the average travel speed $v$ of $\vp$. Assume that we have a set of historical trips, $\sP$, that are computed from trajectories. 

A {\bf trip} $\vp$ is defined as a tuple $\displaystyle \vp=(o, d, t, l, \tau)$, where $o, d $ denote an origin and a destination, $t$ is a departure time, $\displaystyle l$ represents the trip distance, and $\tau$ is the travel time of the trip. Given $\vp.l$ and $\vp.\tau$, we derive the average travel speed $v$ of $\vp$. We use $\sP$ to denote a set of historical trips. %that are computed from trajectories. 

To capture the time-dependent traffic, we partition the time domain $\mathit{TI}$ of interest, e.g., a day, into a number of time intervals, e.g., 96 15-min intervals.
For each time interval $T_i\in\mathit{TI}$, we obtain the set of historical trips $\sP_{T_i}$ from $\sP$ whose departure times belong to time interval $T_i$, i.e., $\sP_{T_i} = \{\vp_\vi | \vp_\vi.t \in T_i \wedge \vp_\vi \in \sP \}$.

We further partition a city into $M$ regions $\sV = \{\sV_1, \cdots, \sV_M\}$. An {\bf Origin-Destination (OD) pair} is defined as a pair of regions $(\sV_o, \sV_d)$ where $1\le o, d, \le M$.
%
%geographically such that we can have $M$ subsets of vertices, i.e., $\sV = \{\sV_1, \cdots, \sV_M\}$. 
%

Given a time interval $T_i$, two regions $\sV_o$ and $\sV_d$, we obtain a trip set $\sP_{T_i,\sV_o, \sV_d} = \{\vp_i | \vp_i.o \in \sV_o \wedge \vp_i.d \in \sV_d\wedge \vp_i.t \in T_i \}$, meaning that each trip in $\sP_{T_i,\sV_o, \sV_d}$ starts from region $\sV_o$, at a time in interval $T_i$, and ends at region $\sV_d$.

Next, we construct an equi-width histogram $\sH_{T_i,\sV_o, \sV_d}$ to record the stochastic speed of trips in $\sP_{T_i,\sV_o, \sV_d}$.
In particular, an equi-width histogram is a set of $K$ bucket-probability pairs, i.e., $\sH_{T_i,\sV_o, \sV_d} = \{(b_j, \mathit{pr}_j)\}$. A bucket $b_j = [v_s,v_e)$ represents the speed range from $v_s$ to $v_e$, and all buckets have the same range size. Probability $\mathit{pr}_j$ is the probability that the average speed of a trip falls into the range $b_j$.
For example, the speed histogram $\{([0, 20), 0.5)$, $([20, 40), 0.3)$, $([40, 60), 0.2)\}$ for $\sP_{T_i,\sV_o, \sV_d}$ means that the probabilities that the average speed (km/h) of a trip in $\sP_{T_i,\sV_o, \sV_d}$ falls into $[0, 20)$, $[20, 40)$, and $[40, 60)$ are 0.5, 0.3, and 0.2, respectively.

\begin{definition}
Given a time interval $T_i$, an {\bf OD stochastic speed tensor} is defined as a matrix $\tM^{(i)}\in \sR^{N\times N' \times K}$, where the first and second dimensions range over the origin and destination regions, respectively, and the third dimension ranges over the stochastic speeds.
For generality, the origin and destination regions can be the same or can be different; thus the first and second dimensions have $N$ and $N'$ instances, respectively.
The third dimension defines $K$ speed buckets.

$M_{o, d, k}$ represents the element $(o,d,k)$ of tensor $\tM^{(i)}\in \sR^{N\times N' \times K}$ and represents the probability of trips in $\sP_{T_i,\sV_o, \sV_d}$ traveling at an average speed that falls into the $k$-th bucket. % defined by the histogram $\sH_{T_i,\sV_o, \sV_d}$.
\end{definition}

Following the example in Figure~\ref{fig:mot}(b), given a time interval $T_i$, for origin region 7 and destination region 8, we obtain a stochastic speed of trips as a histogram, in which the first bucket records that the probability of trips, starting at region 7 during time interval $T_i$ and ending at region 8, traveling at an average speed of $[5,10)$ is 0.3.

As shown in Figure~\ref{fig:mot}(b), not all cells have a histogram to capture the stochastic speed. Specifically, the cells with question marks have no histograms because no trip records are available for those cells, i.e., $\sP_{T_i,\sV_o, \sV_d}=\emptyset$, so that $\sH_{T_i,\sV_o, \sV_d}=\emptyset$.
We refer to such tensor as {\bf sparse OD stochastic speed tensor}.

Given a time interval $T_i$, we refer to a tensor where each cell has a stochastic speed $\sH_{T_i,\sV_o, \sV_d}$ as a {\bf full OD stochastic speed tensor}. %, as examplified in Figure~\ref{fig:mot}(c) where there are no question marks.

\subsection{Problem Definition}
%Given a set of historical trips, we partition their departure time into $s$ time intervals $T^{(t-s+1)}$, $\ldots$, $T^{(t)}$ and compose $s$ sparse OD stochastic speed tensors $\tM^{(t-s+1)}$, $\ldots$, $\tM^{(t)}$.

Given $s$ sparse OD stochastic speed tensors $\tM^{(t-s+1)}$, $\ldots$, $\tM^{(t)}$ during $s$ historical time intervals $T^{(t-s+1)}$, $\ldots$, $T^{(t)}$, we aim to predict the stochastic speeds for the next $h$ time intervals $T^{(t+1)}$, $\ldots$, $T^{(t+h)}$ in the form of $h$ full OD stochastic speed tensors $\tM^{(t+1)}$, $\ldots$, $\tM^{(t+h)}$ by learning the following function $f$.
\[ f: [\tM^{(t-s+1)},\ldots, \tM^{(t)}] \rightarrow  [\tM^{(t+1)},\ldots, \tM^{(t+h)}] \]

\section{Basic Stochastic Speed Forecasting}

\subsection{Framework and Intuition}
\label{subsec:basicframe}

\begin{figure}[tb]
\begin{center}
   \includegraphics[width=1.0\linewidth]{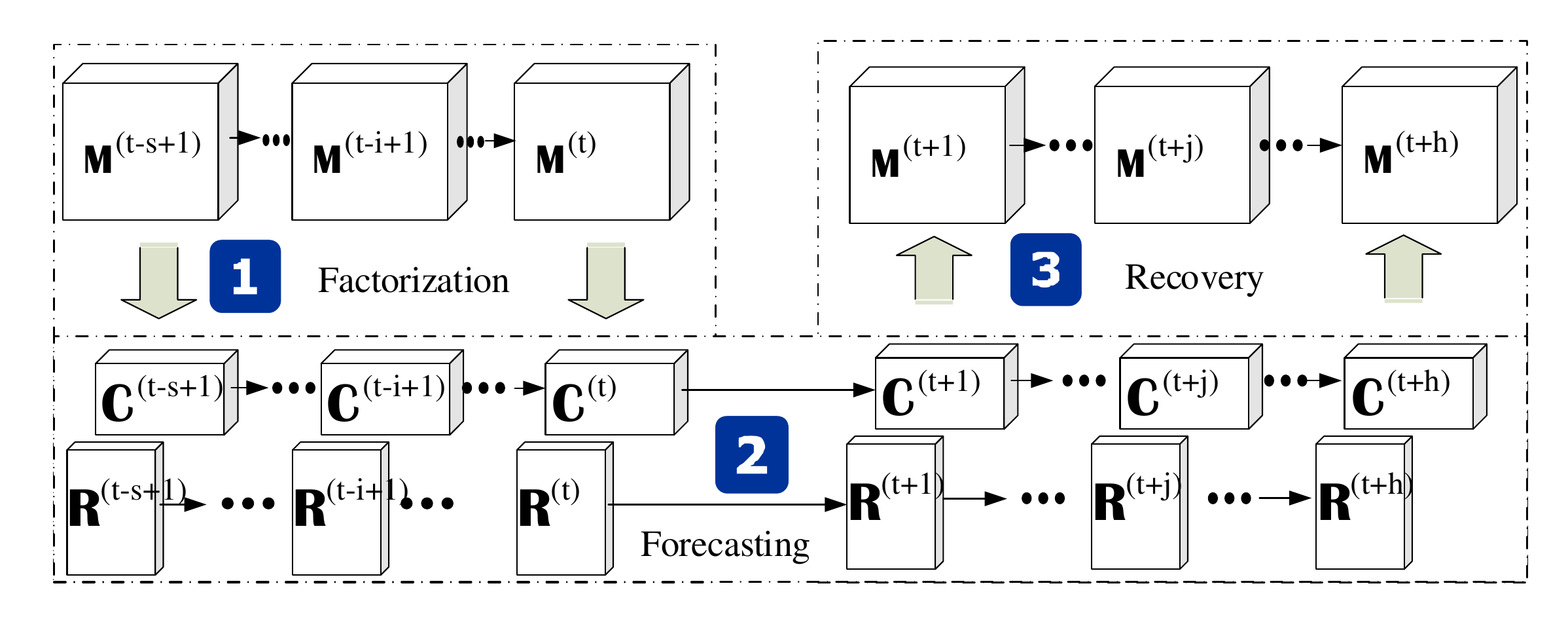}
\end{center}
\caption{Framework Overview.}
\label{fig:basicframe}
\vspace{-10pt}
\end{figure}

Figure~\ref{fig:basicframe} shows the basic framework for forecasting stochastic speeds, which consists of three steps: {\it Factorization}, {\it Forecasting}, and {\it Recovering}.

For the historical time intervals $T^{(t-s+1)}$, $\ldots$, $T^{(t)}$, we have sparse OD stochastic speed tensors $\tM^{(t-s+1)}$, $\ldots$, $\tM^{(t)}$. 
We factorize each stochastic speed tensor $\tM^{(t-i+1)}\in \sR^{N\times N' \times K}$, where $i\in[1,s]$ into two smaller tensors $\tR^{(t-i+1)} \in \sR^{N\times \beta \times K}$ and $\tC^{(t-i+1)}\in \sR^{\beta \times N' \times K}$, where $\beta\ll N, N'$. The aim is to use $\tR^{(t-i+1)}$ and $\tC^{(t-i+1)}$ to approximate $\tM^{(t-i+1)}$.
Here $\tR^{(t-i+1)}$ and $\tC^{(t-i+1)}$ model the correlated features of stochastic speeds among origin regions and among destination regions, respectively. And it is intuitive to assume that stochastic speeds among origin regions and among destination regions share correlated features, as traffic in a region affects the traffic in its nearby regions.
%which allows us to capture the correlated features of stochastic speeds among origin regions and among destination regions, respectively.

The factorization is supported by the intuition underlying low-rank matrix approximation~\cite{DBLP:conf/nips/MontiBB17,srebro2005maximum,srebro2003weighted,marlin2004modeling}. Since $\tM^{(t-i+1)}$ is a sparse tensor, we aim to find a low-rank tensor $\tM'^{(t-i+1)}$ to approximate $\tM^{(t-i+1)}$.
When carrying out the approximation, we assume that the rank of $\tM'^{(t-i+1)}$ is at most $\beta$ and that it can be factorized as $\tM'^{(t-i+1)} = \tR^{(t-i+1)} \times \tC^{(t-i+1)}$.  
Then, the problem of using $\tM'^{(t-i+1)}$ to approximate $\tM^{(t-i+1)}$ can be formulated as the problem of minimizing the following loss function. 
\begin{equation}
\min_{\tR^{(x)}, \tC^{(x)}} ||\tR^{(x)}||_{F}^{2} + ||\tC^{(x)}||_{F}^{2} + \frac{\mu}{2}||\Omega\circ (\tR^{(x)}\tC^{(x)} - \tM^{(x)})||_{F}^{2},
\label{eq:lr_mc}
\end{equation}
where $x=t-i+1$, $||\cdot||_{F}$ denotes the Frobenius norm, 
$\Omega^{(x)}_{o,d,k}=1$ if the element $(o,d)$ of $M^{(x)}$ is not empty, % an non-empty stochastic travel speed, 
% at the $k$-th speed range, 
and $\circ$ is the element-wise tensor multiplication.

Next, we consider $\tR^{(t-s+1)}, \ldots, \tR^{(t)}$ as an input sequence, from which we capture the temporal correlations among the origin regions of $\tM^{(t-s+1)}, \ldots$, $\tM^{(t)}$.
We feed this input sequence into a sequence-to-sequence RNN model~\cite{seq2seq} to \textit{forecast} an output sequence that represents the shared features among the origin regions in the future.   

% and upon which we apply {\it RNN Forecasting} to predict the correlated features among origin regions of the future $h$ time intervals, in the form of tensors $\tR^{(t+1)}, \ldots, \tR^{(t+h)}$. 
%
We apply a similar procedure to $\tC^{(t+1)}, \ldots, \tC^{(t+h)}$ to \textit{forecast} an output sequence that represents the shared features among destination regions in the future.

Finally, we \textit{recover} $\tM^{(t+j)}$ as a full OD stochastic speed tensor from $\tR^{(t+j)}$ and $\tC^{(t+j)}$, $j\in[1,h]$.
Since we obtain the predictions $\tR^{(t+j)}$ and $\tC^{(t+j)}$ from the historical $\tR^{(t-i+1)}$ and $\tC^{(t-i+1)}$, $i\in[1,s]$, the intuition of Equation~\ref{eq:lr_mc} also applies when reconstructing $\tM^{(t+j)}$.

%In {\it Feature Decoding}, we decode each $\vo^{(t+j)}$, with $j\in[1,h]$, into two tensors $\tR^{(t+j)}$ and $\tC^{(t+j)}$. Here, $\tR^{(t+j)}$ and $\tC^{(t+j)}$ capture the correlated features of stochastic speeds among origin regions and among destination regions, respectively. By converting $\vo^{(t+j)}$ into $\tR^{(t+j)}$ and $\tC^{(t+j)}$, we benefit from more accurately recovering $\vo^{(t+j)}$ to $\tM^{(t+j)}$, $j\in[1,h]$, in the {\it Recovering} stage, where $\tM^{(t+j)}$ is a full OD stochastic speed tensor.

%Given a list of observations, $[\tM_{x}^{(t-s+1)},\dots, \tM_{x}^{(t)}]$, we first input each observation into a RNN cell sequentially and thus we have the corresponding RNN states, $[\tZ_{1},\dots, \tZ_{s}]$. In order to capture the temporal dynamics and forecast serveral time step ahead, we adopt \textit{seq2seq} constuction such that we have a sequence of outputs from RNN as row and column factorizations, $[\tR_{1},\dots, \tR_{h}]$ and $[\tC_{1},\dots, \tC_{h}]$, respectively. Finally, the OD forecasts output as $[\tM_{y}^{(t+1)},\dots, \tM_{y}^{(t+h)}]$, where $\tM_{y}^{(t+i)} = \tR_{i}\tC_{i}$.
%
%To forecast the OD stochastic speed effectively, we propose a basic model and an advanced model.
%
%The basic model is using fully-connected layer to fulfill the dimension reduction without considering the spatial dependencies. In addition, the classical seq2seq RNN framework is utilized to capture the temporal dynamics.

%Low rank matrix approximation is a well studied problem for collaborative prediction~\cite{DBLP:conf/nips/MontiBB17,srebro2005maximum,srebro2003weighted,marlin2004modeling}, which aims at searching a low rank approximated matrix $\mX\in \sR^{N\times N'}$ to minimize the loss for a partially observed matrix $\mY\in \sR^{N\times N'}$. Suppose that the matrix $\mX$ with rank at most $m$, it can be factored into $\mX = \mU\mV$, where $\mU \in \sR^{N\times \beta}$ and $\mV\in \sR^{\beta \times N}$. The low rank approximation problem can be formulated as follows~\cite{srebro2005maximum,srebro2003weighted}.
%
%\begin{equation}
%\text{min}_{\mU, \mV} ||\mU||_{F}^{2} + ||\mV||_{F}^{2} + \frac{\mu}{2}||\Omega\circ (\mU\mV - \mY)||_{F}^{2}
%\label{eq:lr_mc}
%\end{equation}
%where $||\cdot||_{F}$ denotes the Frobenius norm, $\Omega$ is the weighted matrix ($\Omega_{i,j} = 1$ iff $\mY_{i, j}$ has the observed value), and $\circ$ denotes the element-wise matrix multiplication.

%
%\subsection{Forecast w/o Spatial Dependency}
%\subsubsection{Input Feature}
\subsection{Factorization}
\label{subsec:factorization}
Given an input sparse OD stochastic tensors $\tM^{(t-i+1)}\in \sR^{N\times N'\times K}$ for intervals $T^{(t-i+1)}$, where $i\in[1,s]$, we proceed to describe the method for factorizing $\tM^{(t-i+1)}$ into $\tR^{(t-i+1)}$ and $\tC^{(t-i+1)}$, which are able to capture the correlated features of stochastic speed among origin and destination regions, respectively.
%+++l or alpha+++
We first flatten $\tM^{(t-i+1)}$ into a vector $\vf^{(t-i+1)}\in \sR^l$, where $l=N \cdot N'\cdot K$, from which 
we generate two small factorization vectors, $\vc^{(t-i+1)}\in \sR^{N'\cdot K \cdot \beta}$ and $\vr^{(t-i+1)}\in \sR^{N\cdot K \cdot \beta}$ via a fully-connected neural network layer (FC layer).%, which capture the correlated features of stochastic speed among origin regions and among destination regions, respectively.
\begin{equation}
\vr^{(t-i+1)} = \text{relu}(\mF_\vr \times \vf^{(t-i+1)} + \vb_r) %\text{ and}
\end{equation}
\begin{equation}
\vc^{(t-i+1)} = \text{relu}(\mF_\vc \times \vf^{(t-i+1)} + \vb_c)
\end{equation}
Here $\mF_\vr \in \sR^{(N\cdot K \cdot \beta) \times l}$  and $\mF_\vc \in \sR^{(N'\cdot K \cdot \beta) \times l}$ are parameter matrices, where $\beta$ is a hyper-parameter to be set; $\vb_r \in \sR^{(N\cdot K \cdot \beta)}$ and $\vb_c\in \sR^{(N'\cdot K \cdot \beta)}$ are bias vectors; and $\text{relu}(\cdot)$ is the relu activation function. %

Next, we reorganize the factorization vectors $\vr^{(t-i+1)}$ and $\vc^{(t-i+1)}$  into factorization tensors $\tR^{(t-i+1)} \in \sR^{N\times \beta \times K}$ and $\tC^{(t-i+1)}\in \sR^{\beta \times N' \times K}$, respectively. %Therefore, we have two factorization tensor sequences $\{\tR^{(t-s+1)},\dots, \tR^{(t)}\}$ and $\{\tC^{(t-s+1)},\dots, \tC^{(t)}\}$ .

%Then we multiply a matrix weight $\mT_E \in \sR^{ l \times \alpha}$ with $\vf$, and add up with a biased vector $\vb_E \in \sR^{l}$. Finally, we apply a non-linear function $\epsilon(\cdot)$, displayed as follows.
%\begin{equation}
%\vu^{(t-i+1)} =\epsilon(\mT_E \times \vf^{(t-i+1)} + \vb_E), i\in[1,s].
%\end{equation}
%where, $\epsilon(\cdot)$ is called activate function, $\mT_I \in \sR^{ l \times N\cdot N'\cdot m}$ is a weight matrix and $\vb_I \in \sR^{l}$ is the bias vector for FC layer.

%\subsubsection{RNN Architecture}
\subsection{Forecasting}
Given historical time intervals $T^{(t-s+1)},\ldots,T^{(t)}$, we learn the temporal correlations of $\tM^{(t-s+1)},\ldots,\tM^{(t)}$ from the temporal correlations among origin regions $\tR^{(t-s+1)},\dots, \tR^{(t)}$ and the temporal correlation among destination regions $\tC^{(t-s+1)},\dots, \tC^{(t)}$.

Based on $\tR^{(t-s+1)},\dots, \tR^{(t)}$, we use a sequence-to-sequence RNN model~\cite{seq2seq} to forecast $\widehat{\tR}^{(t+1)},\dots, \widehat{\tR}^{(t+h)}$ for the future time intervals $T_{t+1}$, $\ldots, T_{t+h}$. %  from $\tR^{(t-s+1)},\dots, \tR^{(t)}$.
In particular, we apply Gated Recurrent Units~(GRUs) in the RNN architecture, since these can capture temporal correlations well by using gate units well and also offer high efficiency~\cite{cho2014learning,chung2014empirical}.
%
%Further, we use \textit{seq2seq}~\cite{seq2seq} construction to obtain prediction output. 
The process is presented as follows. % as $\vo^{(t+1)}$, $\ldots, \vo^{(t+h)}$ from input $\vu^{(t-s+1)}$, $\ldots, \vu^{(t)}$, presented as follow.
\begin{equation}
\label{eq:gru_past}
\widehat{\tR}^{(t+1)},\dots, \widehat{\tR}^{(t+h)} = \mathit{seq2seqGRU}(\tR^{(t-s+1)},\dots, \tR^{(t)}).
\end{equation}

A similar procedure is applied to obtain $\widehat{\tC}^{(t+1)},\dots, \widehat{\tC}^{(t+h)}$ from $\tC^{(t-s+1)},\dots, \tC^{(t)}$.

\subsection{Recovery}
\label{sect:recovering}
Given predicted tensors $\widehat{\tR}^{(t+j)} \in \sR^{N\times \beta \times K}$ and $\widehat{\tC}^{(t+j)}\in \sR^{\beta\times N' \times K}$ for a future time interval $T^{(t+j)}$, with $j\in[1,h]$, we proceed to describe how to transform $\widehat{\tR}^{(t+j)}$ and $\widehat{\tC}^{(t+j)}$  into a full OD stochastic speed tensor $\tM^{(t+j)}\in \sR^{N\times N' \times K}$.

First, we slice each of $\widehat{\tR}^{(t+j)}$ and $\widehat{\tC}^{(t+j)}$ by the speed bucket dimension into $K$ matrices. Specifically, we have 
%
%$\Mathit{slice}(\tR^{(t+j)}) = \{\tR_{:,:,1}^{(t+j)} \cdots, \tR_{:,:,K}^{(t+j)} \}$ and $\Mathit{slice}(\tC^{(t+j)}) = \{\tC_{:,:,1}^{(t+j)} \cdots, \tC_{:,:,K}^{(t+j)} \}$, where $\tR_{:,:,k}^{(t+j)} \in \sR^{N\times \beta}$ and $\tC_{:,:,k}^{(t+j)} \in \sR^{N'\times \beta}$, $k\in \{1, \cdots, K\}$.
%
$\mathit{slice}(\widehat{\tR}^{(t+j)}) = \{\widehat{\tR}_{:,:,1}^{(t+j)} \cdots, \widehat{\tR}_{:,:,K}^{(t+j)} \}$ and $\mathit{slice}(\widehat{\tC}^{(t+j)}) = \{\widehat{\tC}_{:,:,1}^{(t+j)} \cdots, \tC_{:,:,K}^{(t+j)} \}$, where $\widehat{\tR}_{:,:,k}^{(t+j)} \in \sR^{N\times \beta}$ and $\widehat{\tC}_{:,:,k}^{(t+j)} \in \sR^{\beta\times N'}$, $k\in [1,K]$.

Next, we conduct a matrix multiplication as follows.
\begin{equation}
\widetilde{\mM}_k^{(t+j)} = \widehat{\tR}_{:,:,k}^{(t+j)} \times (\widehat{\tC}_{:,:,k}^{(t+j)}),
\end{equation}
where $\widetilde{\mM}_k^{(t+j)}\in \sR^{N\times N'}$, $k\in [1,K]$.

Finally, we are able to construct a tenor $\widetilde{\tM}^{(t+j)} \in  \sR^{N\times N' \times K}$ by combining a total of $K$ matrices, i.e., $ \widetilde{\tM}_{:, :, k}^{(t+j)} = \widetilde{\mM}_k^{(t+j)}$, $k\in [1,K]$. Now, $\widetilde{\tM}^{(t+j)}$ is a full tensor where each element has a value.

A histogram $ \widehat{\tM}_{o, d, :}^{(t+j)}\in \sR^{1\times K}$ %contained in a full OD stochastic speed tensor $\widehat{\tM}^{(t+j)}$, $ \widehat{\tM}_{o, d, :}^{(t+j)}$ 
must meet two requirements to be a meaningful histogram:
(1) $ \widehat{\tM}_{o, d, k}^{(t+j)} \in [0,1]$, $k\in[1,K]$, meaning that the probability of a speed falling into the $k$-th bucket for each OD pair $(o, d)$ must between 0 and 1; and 
(2) $\sum_{k=1}^{K} \widehat{\tM}_{o, d, k}^{(t+j)}  = 1$, meaning that the probability of a speed falling into all $K$ buckets for each $(o, d)$ must equal 1.

To achieve this, we apply a softmax function to normalize values in $\widetilde{\tM}^{(t+j)}$ into $ \widehat{\tM}_{o, d, :}^{(t+j)}$ that satisfies the histogram requirements. 
%
%
%Since we query for the stochastic speed forecasts, e.g., the histograms of the future travel speed. The forecasts need to stratify two constraints: 1)  $\tM_{i, j, k}^{(t+m)} \in [0, 1]$; 2) $\sum_{k=1}^{K} \tM_{i, j, k}^{(t+m)}  = 1$, $\forall i \in \{1, \cdots, N\}$, $ \forall j \in \{1, \cdots, N'\}$ and $\forall k \in \{1, \cdots, K\}$. To this end, a softmax function is applied to the last dimension of $\tM^{(t+m)}$, $\forall m \in \{1, \cdots, h\}$:
\begin{equation}
\widehat{\tM}_{o, d, :}^{(t+j)} = \mathit{softmax}(\widetilde{\tM}_{o, d, :}^{(t+j)}), \forall o\in [1,N] \text{, } \forall d \in [1,N'].
\end{equation}

Thus, we obtain $h$ meaningful full OD stochastic speed tensors for the future time intervals $T_{t+1},\dots,T_{t+h}$ as the output of the recovery process: $\widehat{\tM}^{(t+1)},\dots, \widehat{\tM}^{(t+h)}$.

\subsection{Loss Function}
\label{sect:loss1}
%According to Equation~\ref{eq:lr_mc}, the loss function can be constructed by minimizing on the factorizations directly. However, the softmax operation in the output layer makes it not a ``real'' matrix factorization. Hence, 
%
The loss function is defined as the error between the recovered future tensor and the ground-truth future tensor. 
%we present the loss function as follows.
\begin{align}
\begin{split}
\ell(\mF, \vb) = \sum_{j=1}^{h}[&\lambda||\widehat{\tR}^{(t+j)}||_{F}^{2} + \lambda||\widehat{\tC}^{(t+j)}||^{2}_{F} + \\
& ||\Omega^{(t+j)}\circ (\tM^{(t+j)} - \widehat{\tM}^{(t+j)})||_{F}^{2}],
\end{split}
\label{eq:loss_func_BF}
\end{align}
where $\mF$ and $\vb$ represent the training parameters in the framework, and $\lambda$ is a regularization parameter.
Further, $\Omega^{(t+j)}\in \sR^{N\times N' \times K}$ is an indication tensor, where $\Omega^{(t+j)}_{o,d,k}=1$ if the OD pair $(o,d)$ is not empty in the $t+j$'th future interval. Note that although we aim to predict full tensors, the ground truth tensors are sparse, so we compute the errors taking only into account the non-empty elements in the ground truth tensors. 
%
%t the $k$-th historical bucket, 
Next, $\circ$ is element-wise multiplication, $\widehat{\tM}^{(t+j)}$ and $\tM^{(t+j)}$  are predicted and ground truth tensors, respectively,  $||\cdot||_{F}$ is the Frobenius-norm.

\section{Forecast with Spatial Dependency}
\label{sec:forecast_spatial}
To improve forecast accuracy, we proceed to integrate spatial dependency into our framework in two different stages. 
First, in the factorization step, we apply graph convolutional neural networks to perform feature encoding for origin and destination dimensions, respectively.  
Second, in the forecasting step, we integrate graph convolutional with RNNs to capture spatio-temporal correlations. %correlations again temporal correlations. 
%Therefore, the spatial correlations are better preserved to improve the accuracy. %, which will be discussed in Section~\ref{sec:forecast_spatial}.

\subsection{Spatial Factorization}
As in Section~\ref{subsec:factorization}, we aim to factorize tensor $\tM^{(t-i+1)}$ during interval $T_{t-i+1}$, $i\in[1,s]$, into two smaller tensors  $\tC^{(t-i+1)}$ and $\tR^{(t-i+1)}$. % supported by the intuition of Equation~\ref{eq:lr_mc} in this section.
In Section~\ref{subsec:factorization}, $\tM^{(t-i+1)}$ is simply flattened and followed by a fully-connected layer to construct $\tC^{(t-i+1)}$ and $\tR^{(t-i+1)}$. %, in order to capture the correlated features among origin regions and among destination regions, respectively. 
This process does not take spatial correlations among the origin regions and among the destination regions into account, although such correlations are likely to exist.
To accommodate spatial correlations, we first capture spatial correlations among origin and destination regions; then we use the captured spatial correlations to conduct factorization. 

\subsubsection{Spatial Correlation}
\label{sect:sc}
We leverage the notion of a proximity matrix~\cite{li2018dcrnn_traffic} to capture spatial correlations. 
We proceed to present the idea using origin regions as an example, which also applies to destination regions in a similar manner.

Given $\tM^{(t-i+1)}\in \sR^{N\times N' \times K}$, we have $N$ origin regions, from which we build an adjacency matrix $\mA\in\sR^{N\times N}$ to show  region connections.
Specifically, $\mA_{u,v}=1$ means that regions $\sV_u$ and $\sV_v$ are adjacent; otherwise, $\mA_{u,v}=0$. 

We construct a weighted proximity matrix $\mW^{(\alpha, \sigma)}\in \sR^{N\times N}$ from $\mA$ that describes the proximity between regions $\sV_u$ and $\sV_v$ and is parameterized by adjacency hops $\alpha$ and standard deviation $\sigma$.
Specifically, if $\sV_v$ can be reached from $\sV_u$ in $\alpha$ adjacency hops using $\mA$, $\mW^{(\alpha, \sigma)}_{u, v} = e^{-x^2/\sigma^2}$, where $x$ is the distance between the centroid of $\sV_u$ and $\sV_v$; otherwise $\mW^{(\alpha, \sigma)}_{u, v}=0$. In the experiments, we study the effect of $\alpha$ and $\sigma$ (see Section~\ref{sect:exp_pm}). 
The proximity matrix $\mW^{(\alpha, \sigma)}$ is symmetric and non-negative.

The adjacency matrices for the source regions and destination regions may be different or the same. Consider two scenarios. First, we use OD matrices to model the travel costs within a city. In this case, the source regions and the destination regions are the same, and thus the two adjcency matrices are the same. Second, we may use OD matrices to model the travel costs between two different cities. Then, the source regions and the destination regions are in different cities. Thus, we need two different adjacency matrices. 
To avoid confusion, we use $\mW$ and $\mW^\prime$ represent the adjacency matrices for source regions and destination regions, respectively. 

\subsubsection{Factorization}

%Since the relationship among regions in RoIs can be defined in a graph $\gG$, we come up with a better fatorization approach, i.e. factorizing $\tM^{(t-i+1)}$ into $\tR^{t-i+1}$ and $\tC^{(t-i+1)}$ by taking advantage of the spatial correlations on both 1st and 2nd dimension of $\tM^{(t-i+1)}$, respectively. Figure~\ref{fig:feg} shows a vivid example of such operation for $\tR^{(t-i+1)}$. Similar process will be applied to obtain $\tC^{(t-i+1)}$.

We proceed to show the factorization procedure. Specifically, we show how to obtain $\tR^{(t-i+1)}$ from $\tM^{(t-i+1)}$. The same procedure can be applied to obtain $\tC^{(t-i+1)}$.

\begin{figure}[!ht]
\begin{center}
   \includegraphics[width=1.0\linewidth]{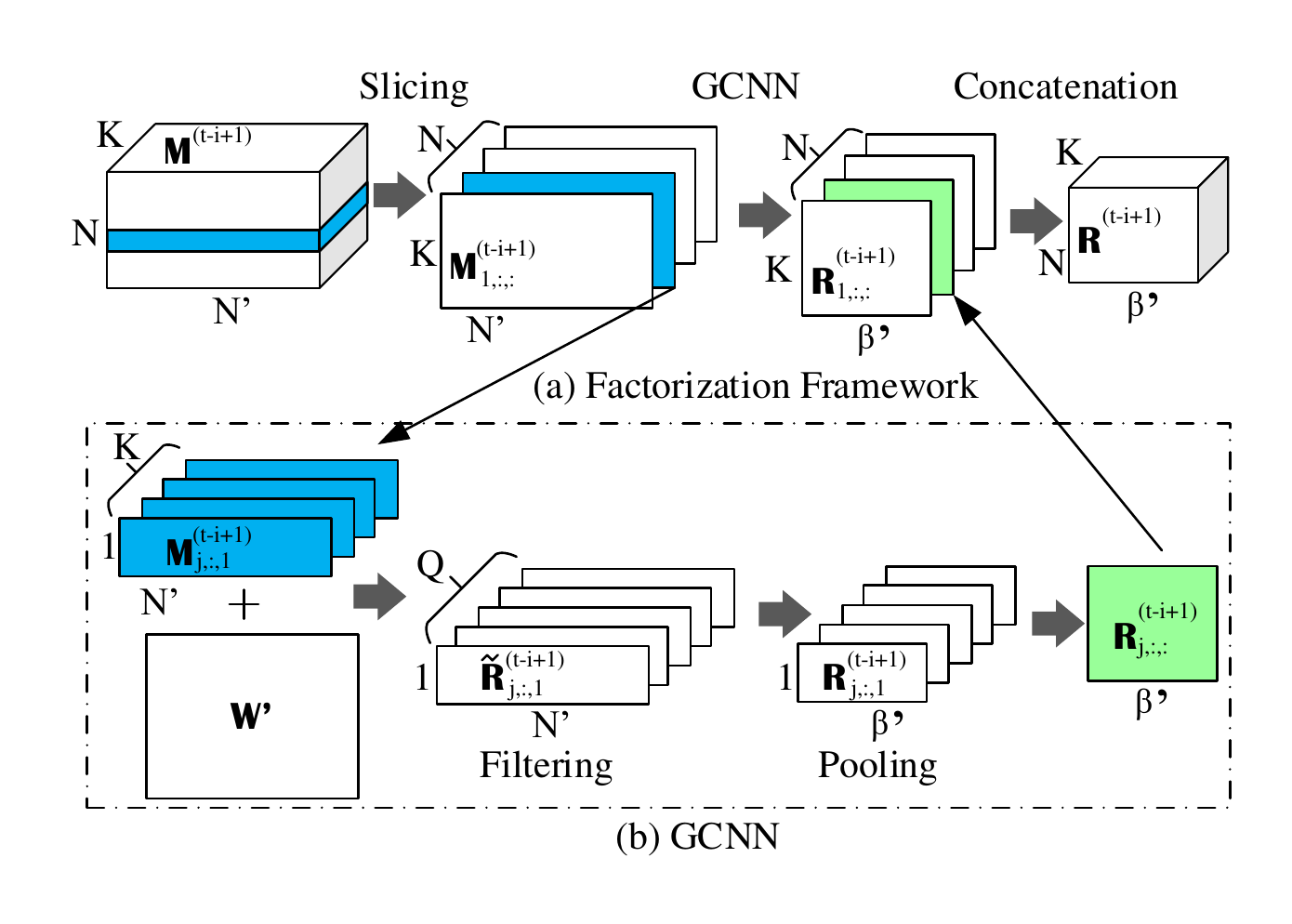}
\end{center}
\caption{Spatial Factorization for $\mR$}
\label{fig:feg}
\end{figure}

As shown in Figure~\ref{fig:feg}(a), we first slice $\tM^{(t-i+1)}\in\sR^{N\times N' \times K}$ by the origin region dimension into $N$ matrices, i.e., $\mathit{slice}(\tM^{(t-i+1)}) = [\tM_{1,:, :}^{(t-i+1)}, \cdots, \tM_{N, :, :}^{(t-i+1)}]$. Each of the sliced matrix is then applied with a GCNN operation. 
Accordingly, we obtain the GCNN output as $[\tR_{1,:, :}^{(t-i+1)}$, $\cdots, \tR_{N, :, :}^{(t-i+1)}]$. We then concatenate this to obtain $\tR^{(t-i+1)}\in\sR^{N\times \beta' \times K}$.

%In Figure~\ref{fig:feg}, we first slice $\tM^{(t-i+1)}$ by the 1st dimension into $N$ matrices, i.e., $slice(\tM^{(t-i+1)}) = \{\tM_{1,:, :}^{(t-i+1)}, \cdots, \tM_{N, :, :}^{(t-i+1)}\}$, which will be applied with same GCNN operation, separatively.

Figure~\ref{fig:feg}(b) shows a GCNN operation on a sliced matrix $\tM_{j, :, :}^{(t-i+1)}$ $j\in[1, N]$, which transforms $\tM_{j,:, :}^{(t-i+1)}\in\sR^{K \times N'}$ into $\tR_{j,:, :}^{(t-i+1)}\in\sR^{K \times \beta'}$ via {\it Filtering} and {\it Pooling}. 

%that learns $\beta'$ spatial correlated features among $N'$ destination regions. In particular, we show how a GCNN transforms each $\tM_{j,:, :}^{(t-i+1)}\in\sR^{N' \times K}$, $j\in[1,N]$, into $\tR_{j,:, :}^{(t-i+1)}\in\sR^{\beta' \times K}$ via {\it Filtering} and {\it Pooling}.

\noindent{\bf Filtering: }
%We take $j$-th slice, $\tM_{j,:, :}^{(t-i+1)} \in \sR^{(N'\times K)}$, as an example which is filled with blue color in Figure~\ref{fig:feg} and this GCNN process is also shown in the bounding box below. 
%For each $\tM_{j,:, :}^{(t-i+1)}\in\sR^{N' \times K}$, $j\in[1,N]$, we further slice it into $K$ vectors $[\tM_{j,:, 1}^{(t-i+1)}, \cdots, \tM_{j,:, K}^{(t-i+1)}]$, where vector $\tM_{j,:, k}^{(t-i+1)}\in \sR^{N'}$, $k\in[1,K]$, presents the probability of speeds falling at the $k$-th speed bucket when traveling from the origin region $\sV_j$ to all destination regions.
%
Given $\tM_{j,:, :}^{(t-i+1)}\in\sR^{K \times N'}$, we apply $Q$ graph convolutional filters, which take into account the destination region adjacency matrix $\mW$, to generate $\widetilde{\tR}_{j, :, :}^{(t-i+1)} \in \sR^{N'\times Q}$ that captures the correlated features among destination regions. 
%

%Specifically, w
We first slice $\tM_{j,:, :}^{(t-i+1)}\in\sR^{K \times N'}$ into $K$ vectors $[\tM_{j,:, 1}^{(t-i+1)}$, $\cdots$, $\tM_{j,:, K}^{(t-i+1)}]$, where vector $\tM_{j,:, k}^{(t-i+1)}\in \sR^{N'}$, $k\in[1,K]$, represents the probability of speeds falling into the $k$-th speed bucket when traveling from origin region $\sV_j$ to all destination regions.
%
%a set of vecor feature maps set with size $K$, i.e., $\{\tM_{j,:, 1}^{(t-i+1)}, \cdots, \tM_{j,:, K}^{(t-i+1)}\}$, where $\tM_{j,:, k}^{(t-i+1)}\in \sR^{N'}$ which is a graph signal. 

Next, we use a specific graph convolutional filter, namely Cheby-Net~\cite{defferrard_convolutional_2016}, due to its high accuracy and efficiency, on each vector $\tM_{j,:, k}^{(t-i+1)}$.
%
%, to learn correlated features among destination regions as $\widehat{\tR}_{j, :, :}^{(t-i+1)} \in \sR^{N'\times Q}$, where $Q$ is the number of filters, 
%$\widehat{\tR}_{j, :, q}^{(t-i+1)}\in \sR^{N'}$, with $q\in[1,Q]$, 
%from $\tM_{j,:, 1}^{(t-i+1)}, \ldots, \tM_{j,:, K}^{(t-i+1)}$.
%
%For each bucket $k\in[1,K]$, 
Specifically, before conducting actual convolutions, we compute $\mT_k^{(t-i+1)} = [\vt_1, \vt_2, \ldots, \vt_{S}]$, where $\vt_s\in\sR^{N'}$, $s\in[1,S]$, from $\tM_{j,:, k}^{(t-i+1)}$. 
Here, $\vt_{1}=\tM_{j,:, k}^{(t-i+1)}$, $\vt_{2}=\hat{\mL}\times \tM_{j,:, k}^{(t-i+1)}$, and $T_{s}=2\hat{\mL}\times T_{s-1} - T_{s-2}$ when $s>2$, where $\hat{\mL} = 2\mL/\lambda_{max} - \mI$ is a scaled Laplacian matrix and where $\mL = \mD - \mW^\prime$ is the Laplacian matrix and $\lambda_{max}$ is the maximum eigenvalue of $\mL$. 
Here, we use destination adjacency matrix $\mW^\prime$ because $\tM_{j,:, k}^{(t-i+1)}$ represents the speed from source region $j$ to all destination regions and we use $\mW^\prime$ to capture the spatial correlation among destination regions. 
%Following an existing approximation method~\cite{defferrard_convolutional_2016}, we use $\lambda_{max} = 2$ as the default.
%
After the whole computation, we get $\mT_k\in\sR^{N'\times S}$ as the encoded features for the $k$-th bucket while considering the spatial correlations among destination regions. 

%For total $K$ buckets, we sum up matrices, i.e., $\mT = \mT_1 +, \ldots, + \mT_K$, and $\mT\in\sR^{N'\times S}$ as encoded features among destination regions for all buckets. 

%To conduct graph convolution, 

Then we proceed to apply $Q$ filters to $\mT_k$. Each filter is a vector $\tG_q\in\sR^{S}$, where $q\in[1,Q]$. 
We apply each filter to all $\{\mT_k^{(t-i+1)}\}$, $\forall k \in [1, K]$, and then the sum is used as the output of the filter. 
\begin{equation}
\label{eq:graphconv}
\widetilde{\tR}_{j, :, q}^{(t-i+1)}=\tG_q  \otimes \tM_{j,:, :}^{(t-i+1)}  = \sum_{k=1}^{K}\big( \epsilon(\mT_k^{(t-i+1)} \times \tG_q + \vb_{q}),
\end{equation}
where $\otimes$ is the Cheby-Net graph convolution operation, $\vb_{q}\in \sR^{N'}$ is a bias vector, and $\epsilon(\cdot)$ is a non-linear activate function.

Finally, we arrange the results obtained from all $Q$ filters as $\widetilde{\tR}_{j, :, :}^{(t-i+1)} = [\widetilde{\tR}_{j, :, 1}^{(t-i+1)} $, $\ldots$, $\widetilde{\tR}_{j, :, Q}^{(t-i+1)}]$, where $\widetilde{\tR}_{j, :, :}^{(t-i+1)} \in \sR^{Q\times N'}$.

%%
%We apply each filter to learn representative features as $\widehat{\tR}_{j, :, q}^{(t-i+1)}\in\sR^{N'}$ from the encode features as $ \widehat{\tR}_{j, :, q}^{(t-i+1)} = \mT \times \tG_q$.
%%
%We arrange results from all $Q$ filters as $\widetilde{\tR}_{j, :, :}^{(t-i+1)} = [\widehat{\tR}_{j, :, 1}, \ldots, \widehat{\tR}_{j, :, Q}]$.
%%
%We denote the above graph filtering process as: 
%\begin{equation}
%\label{eq:gcnn}
%\widetilde{\tR}_{j, :, :}^{(t-i+1)} = \tG  \otimes \tM_{j,:,:}^{(t-i+1)}.
%\end{equation}
%
%where $\mG\in \sR^{Q\times S}$ is the weight parameter tensor for GCNN ($\mG_{q, :}$ is called one GCNN filter and $S$ is the filter size), $Q$ is the number of filters, $T_{0}(\tM_{j,:, k}^{(t-i+1)})=\tM_{j,:, k}^{(t-i+1)}$, $T_{1}(\tM_{j,:, k}^{(t-i+1)})=\hat{\mL}\times \tM_{j,:, k}^{(t-i+1)}$, and $T_{s}(\tM_{j,:, k}^{(t-i+1)})=2\hat{\mL}\times T_{s-1}(\tM_{j,:, k}^{(t-i+1)}) - T_{s-2}(\tM_{j,:, k}^{(t-i+1)})$ when $s \geq 2$, where $\hat{\mL} = 2\mL/\lambda_{max} - \mI$ is a scaled Laplacian Matrix where $\mL = \mD - \mW$ is the laplacian matrix, $\lambda_{max}$ is the maximum eigenvalue of $\mL$, in this paper we take $\lambda_{max}= 2$ for simplicity.

%We apply a non-linear activation function to $\widetilde{\tR}_{j, :, :}^{(t-i+1)} \in \sR^{N'\times Q}$.
%\begin{equation}
%\widehat{\tR}_{j, :, :}^{(t-i+1)} = \epsilon(\widetilde{\tR}_{j, :, :}^{(t-i+1)} + \vb_{\widehat{\tR}_{j, :, :}})
%\end{equation}
%where $\vb_{\widehat{\tR}_{j, :, :}}\in \sR^{N'}$ is the bias vector, $\epsilon(\cdot)$ is a non-linear activate function, e.g., sigmoid, tanh, or relu etc.

\noindent\textbf{Pooling: }
To further condense the features and to construct the final factorizations, we apply geometrical pooling~\cite{defferrard_convolutional_2016} to $\widetilde{\tR}_{j, :, :}^{(t-i+1)}$ over the destination region dimension to obtain $\tR_{j, :, :}^{(t-i+1)} \in \sR^{Q \times \beta'}$, where $\beta'=\frac{N'}{p}$ and $p$ are the pooling and stride size, repectively. This process is shown as follows.
\begin{equation}
\tR_{j, :, :}^{(t-i+1)}= \text{P}(\widetilde{\tR}_{j, :, :}^{(t-i+1)}),
\end{equation}
where $\text{P}(\cdot)$ is the pooling function that can be either max pooling or average pooling.

Since the pooling operation requires meaningful neighborhood relationships, we identify spatial clusters of destination regions. For example, in Figure~\ref{fig:poly_rep}, if we use the order of ascending region ids, i.e., (1, 2, 3, 4, 5, 6, 7, 8) to conduct pooling with a pooling size of 2, then regions 3 and 4 are pooled together. However, regions 3 and 4 are not neighbors, so this procedure may yield inferior features that may in turn yield undesired results. 
Instead, if we identify clusters of regions, we are able to produce a new order, e.g., (6, 1, 2, 3, 5, 4, 7, 8). When again using a pooling size of 2, each pool contains neighbouring regions.

The GCNN process, including filtering and pooling, is repeated several times with different numbers of filters $Q$ and pooling stride size $p$. Eventually, we set $Q=K$ and get $\tR_{j, :, :}^{(t-i+1)} \in \sR^{\beta'\times K}$.

As shown in Figure~\ref{fig:feg}(b), the last operation is concatenation. 
We slice $\tM^{(t-i+1)}$ by the origin region dimension into $N$ matrices $[\tM_{1, :, :}^{(t-i+1)}, \ldots, \tM_{N, :, :}^{(t-i+1)}]$ and apply GCNN to each of them to obtain $[\tR_{1, :, :}^{(t-i+1)}, \ldots, \tR_{N, :, :}^{(t-i+1)}]$, where each $\tR_{j, :, :}^{(t-i+1)} \in \sR^{\beta' \times K}$. We then concatenate the $\tR_{j, :, :}^{(t-i+1)}$, $j\in[1,N]$, to obtain $\tR^{(t-i+1)} \in \sR^{N\times \beta' \times K}$. 

The same procedure can be applied to obtain $\tC^{(t-i+1)}$ where we need to change $\mW^\prime$ to $\mW$ when conducting the graph convolution. 

%We conduct factorization to get tensor $\tC^{(t-i+1)} \in \sR^{\beta' \times N' \times K}$ similarly.

%Obviously, we can have more than one GCNN and pooling operations, after which we get the finalized factorization for $j$-th row, e.g., $\tR_{j, :, :}^{(t-i+i)} \in \sR^{\beta \times K}$ which is filled with green color in Figure~\ref{fig:feg}. Finally, we collected all the returuned factorizations, $\{\tR_{1, :, :}^{(t-i+i)}, \tR_{N, :, :}^{(t-i+i)}\}$ , and constructed into a tensor $\tR^{(t-i+i)} \in \sR^{N\times \beta \times K}$. Similarly, we can have a corresponding column factorization tensor $\tC^{(t-i+i)} \in \sR^{\beta \times N' \times K}$.

\subsection{Spatial Forecasting}
To model temporal dynamics while keeping the spatial correlations in RNNs, we combine Cheby-Net based graph convolution with RNNs, yielding CNRNNs. 
Intuitively, we follow the structure of gated recurrent units while replacing the traditional fully connected layer by a Cheby-Net based graph convolution layer. 
Separate CNRNNs are employed to process $\tR^{(t)}$ and $\tC^{(t)}$. 

Taking the source region dimension as an example, a CNRNN takes as input $\tR^{(t)}$ at time interval $T^{(t)}$, and it predicts $\widehat{\tR}^{(t+1)}$ for the future time interval $T^{(t+1)}$. This procedure is formulated as follows. 
\begin{align}
& {\tS}^{(t+1)} = \sigma(\tG_{\tS} \otimes [\tH^{(t)}:{\tR}^{(t)}] + \vb_{\tS})  \label{eq:input} \\
& {\tU}^{(t+1)} = \sigma(\tG_{\tU} \otimes [\tH^{(t)}: \tR^{(t)}] + \vb_{{\tU}})\\
& {\tH}^{(t+1)} =  \tanh(\tG_{\tH} \otimes [ \tR^{(t)}: ({\tS}^{(t+1)}\circ\tH^{(t)})] + \vb_{{\tH}})\\
& \widehat{\tR}^{(t+1)} = {\tU}^{(t+1)} \circ {\tR}^{(t)} + (1 - {\tU}^{(t+1)}) \circ  {\tH}^{(t+1)}
\end{align}
where $\tG_{\tS}$, $\tG_{\tU}$, and $\tG_{\tH}$ are graph convolution filters;  
$\tR^{(t)}$ and ${\widehat{\tR}}^{(t+1)}$ are the input and output of a CNRNN cell at time interval $T^{(t)}$, repectively; 
${\tS}^{(t)}$ and ${\tU}^{(t)}$ are the reset and update gates, respectively; %+++ it should be two gates not 3 gates.+++
$\otimes$ denotes the graph convolution which defined in Equation~\ref{eq:graphconv}, and here the graph convolution should take into account source adjacency matrix $\mW$ since $\tR$ captures features of source regions.   
$\circ$ denotes the Hadamard product between two tensors; 
and $\epsilon(\cdot)$, $\sigma(\cdot)$, and $\tanh(\cdot)$ are non-linear activation functions.

When applying CNRNN to predict ${\widehat{\tC}}^{(t+1)}$, we need to change $\mW$ to $\mW^\prime$ when conducting the graph convolution as $\tR$ captures features of destination regions. 

%where $\tR^{t}$, ${\widehat{\tR}}^{t}$ denote the input and output of CNRNN cell at time t; ${\tS}^{t}$ and ${\tU}^{t}$ are the reset and update gate, respectively; $\otimes$ denotes the graph filtering which is the same as in Equation~\ref{eq:gcnn}; $\tG_{\cdot}$ denote the parameter matrices for GCNN filters; $\circ$ denotes the Hadamard product between two tensors; 

%$\epsilon(\cdot)$ is the non-linear activate function. 

%Similarly, we have $\widehat{\tC}^{(t+1)}$ as the learn features among origin regions for time interval $T_{t+1}$. 
%
Given predicted factorization tensors $[\widehat{\tR}^{(t+1)},\dots, \widehat{\tR}^{(t+h)}]$ and $[\widehat{\tC}^{(t+1)},\dots, \widehat{\tC}^{(t+h)}]$, we apply the same recovery operation introduced in Section~\ref{sect:recovering} to obtain $h$  full OD stochastic speed tensors for the future time intervals $T^{(t+1)},\ldots,T^{(t+h)}$ as the recovery output: $\widehat{\tM}^{(t+1)},\dots, \widehat{\tM}^{(t+h)}$.

\subsection{Loss Function}
%\subsection{Geometric Matrix Completion}
%When there exist the geometric structures embedded on both row and column of the matrix, the forecast problem can be boiled down to be geometric matrix approximation~\cite{benzi2016song,kalofolias2014matrix,rao2015collaborative,ma2011recommender,DBLP:conf/nips/MontiBB17}, which can be reformulated as follows.
%
%\begin{equation}
%\text{min}_{\mX} ||\mX||_{\mW_r}^{2} + ||\mX||_{\mW_c}^{2} + \frac{\mu}{2}||\Omega\circ (\mX - \mY)||_{F}^{2}
%\label{eq:lr_gmc}
%\end{equation}
%where $\mW_r$ and $\mW_c$ are the proximity matrices for graphs on row and column, respectively. $||\mX||_{\mW_r}^{2} = \text{trace}(\mX^T\Delta_r\mX)$ is the Dirichlet norm, where $\Delta_r = \mI - \mD_r^{-1/2}\mW_r\mD_r^{1/2}$ is the normalized graph laplacian matrix and $\mD_r = \text{diag}(\sum_{j\neq i}{\mW_r}(i,j))$ is the degree matrix.
%
%This problem can also be factorized, $\mX = \mU\mV$, such that it has a similar matrix factorized completion problem as follows.
%\begin{equation}
%\text{min}_{\mU, \mV} ||\mU||_{\mW_r}^{2} + ||\mV||_{\mW_c}^{2} + \frac{\mu}{2}||\Omega\circ (\mU\mV - \mY)||_{F}^{2}
%\label{eq:lr_gmfc}
%\end{equation}

%\subsection{Loss Function}
Similar to the construction covered in Section~\ref{sect:loss1}, we present the loss function as follows.
\begin{align}
\begin{split}
\ell(\tG, \vb) = \sum_{i=1}^{h}[&\lambda||\tR^{(t+j)}||_{\mW}^{2} + \lambda||\tC^{(t+j)}||_{\mW}^{2} + \\
& ||\Omega^{(t+j)}\circ (\tM^{(t+j)} - \widehat{\tM}^{(t+j)})||_{F}^{2}]
\end{split}
\label{eq:loss_func}
\end{align}
where $\tG$ and $\vb$ represent the training parameters in the framework (in particular, graph convolutional filters and bias vectors), $||\cdot||_{\mW}^{2}$ is the Dirichlet norm under the proximity matrix $\mW$, $\lambda$ is the regularization parameter for the Dirichlet norm. We use the Dirichlet norm because it takes the adjacency matrix into account---nearby regions should share similar features in the dense tensors $\tR$ and $\tC$. 
Finalltm $\Omega^{(t+j)}\in \sR^{N\times N' \times K}$ is an indication tensor, and $\widehat{\tM}^{(t+j)}$ and $\tM^{(t+j)}$, $j\in[1,h]$,  are the predicted and ground truth tensors, respectively.

\section{Experiments}
We describe the experimental setup and then present the experiments and the findings.

\subsection{Experimental Setup}
\subsubsection{Datasets}
We conduct experiments on two taxi trip datasets to study the effectiveness of the proposal. 

We represent a stochastic speed (m/s) as a histogram with 7 buckets $[0, 3)$, $[3, 6)$, $[6, 9)$, $[9, 12)$, $[12, 15)$, $[15, 18)$, and $[18, \infty)$; and we consider 15-min intervals, thus obtaining 96 15-min intervals per day.

\begin{figure}[t]
	\centering
	\subfigure[NYC Regions]{
		\begin{minipage}[b]{0.22\textwidth}
			\includegraphics[width=1\textwidth]{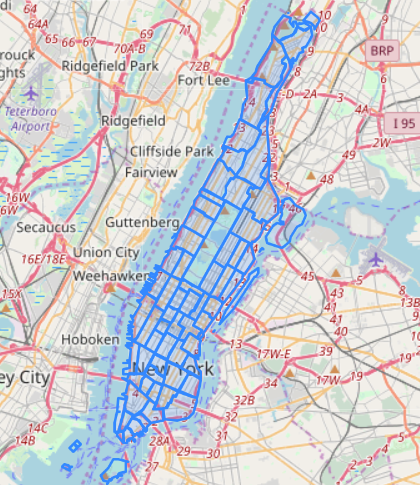}
		\end{minipage}
		\label{fig:nyc_polygon}
	}
	\subfigure[CD Regions]{
		\begin{minipage}[b]{0.22\textwidth}
			\includegraphics[width=1\textwidth]{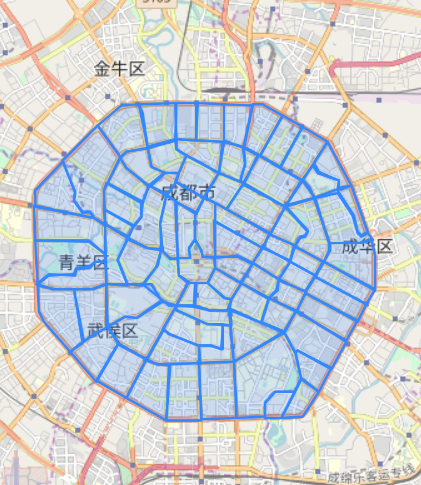}
		\end{minipage}
		\label{fig:cd_polygon}
	}
	\subfigure[NYC Speed]{
		\begin{minipage}[b]{0.22\textwidth}
			\includegraphics[width=1\textwidth]{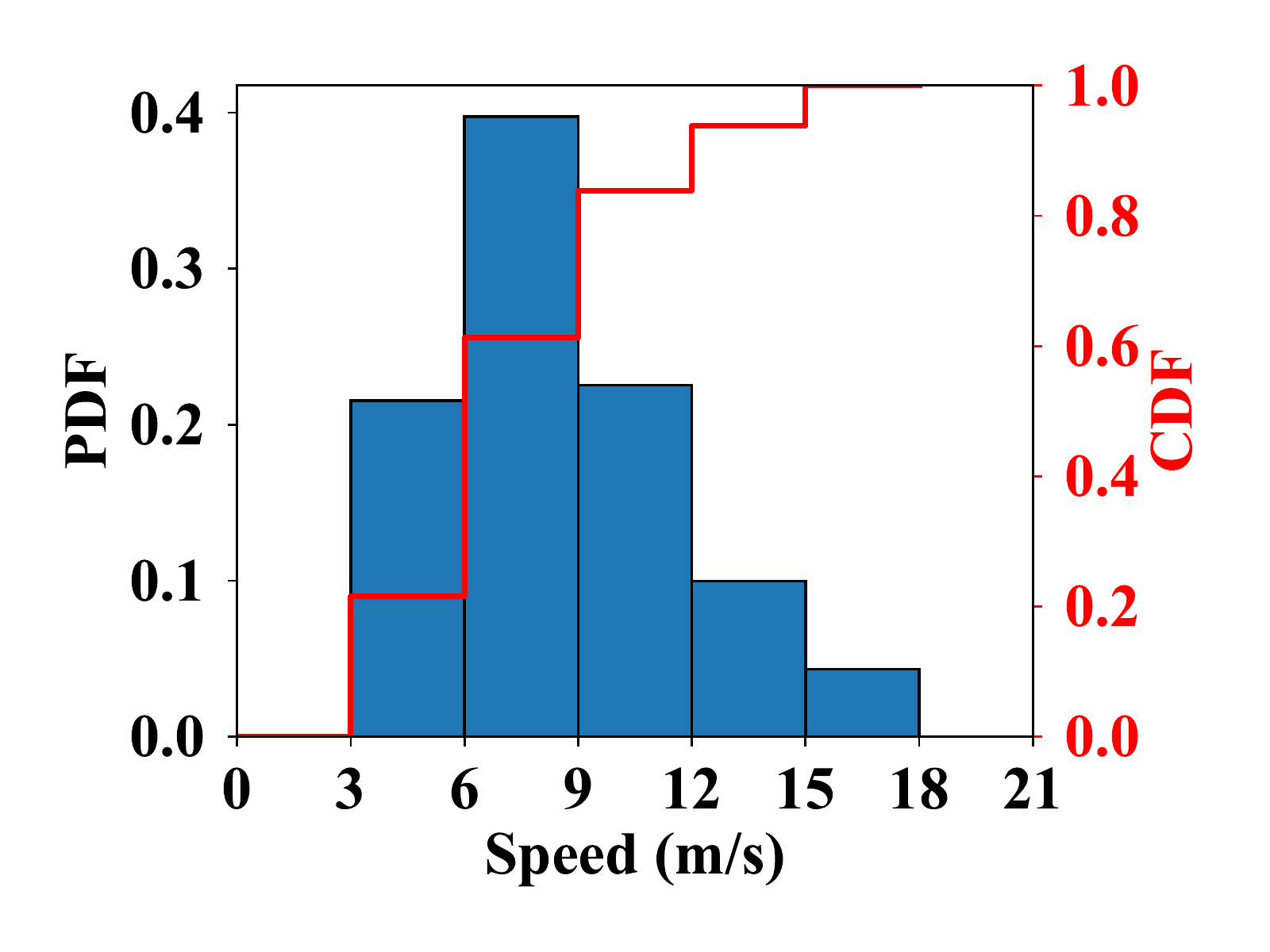}
		\end{minipage}
		\label{fig:nyc_cdf}
	}
	\subfigure[CD Speed]{
		\begin{minipage}[b]{0.22\textwidth}
			\includegraphics[width=1\textwidth]{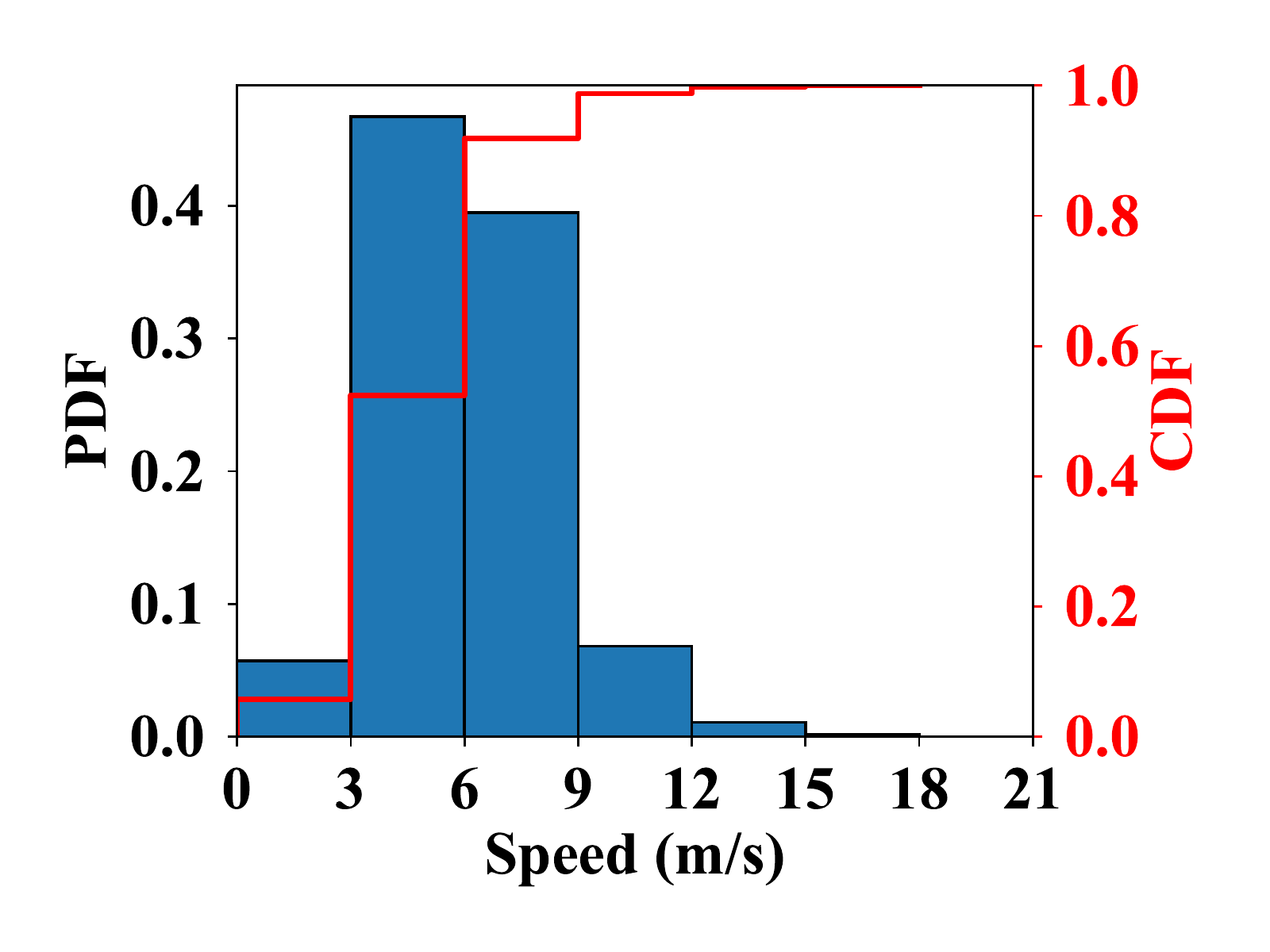}
		\end{minipage}
		\label{fig:cd_cdf}
	}
	\caption{Region Representations of NYC and CD}
	\label{fig:nyc_cd_poly}
\end{figure}

\begin{table}[h]%\normalsize%\small%scriptsize
	\centering
	\begin{tabular}{|p{0.8in}|c|c|}
		\hline
		 & NYC & CD \\ \hline\hline
		\# Trips &14,165,748 &  3,636,845 \\		
		\# Regions & 67 & 79 \\ \hline
		Average Speed & 8.8 m/s & 6.0 m/s\\\hline
		%Time Interval & 15 mins &  15 mins \\\hline

	\end{tabular}
	\caption{Statistics of the Data Sets.}
	\label{tbl:ds}
\end{table}

\noindent
\textbf{New York City Data Set~(NYC): }We use 14 million taxi trips collected from 2013-11-01 to 2013-12-31 from Manhattan, New York City. Each trip consists of a pickup time, a drop off time, a pickup location, a drop off location, and a total distance. 
Manhattan has 67 taxizones\footnote{\url{http://www.nyc.gov/html/tlc/html/about/trip_record_data.shtml}}, each of which is used as a region. The regions are shown in Figure~\ref{fig:nyc_cd_poly}(a). 
%we utlize the taxizone shapefile again to segment Manhattan borough into 67 subregions, as is shown in Figure~\ref{fig:nyc_polygon}. Next, we select the length of time interval to be 15 minutes, i.e., 96 time intervals per day. 
%
The OD stochastic speeds for NYC are represented as an $\sR^{67\times 67 \times 7}$ tensor. %, i.e., 96 tensors per day. 

%Training vs testing data sets: 

%We use 29 million trips collected from 2013-11-01 to 2013-12-31 in New York City. Each trip consists of a pickup time, a drop off time, a pickup location, a drop off location, and a total distance. To get reliable results, we use a subset of trips whose pickup and drop off locations are within Manhattan according to the taxizone shapefile\footnote{\url{http://www.nyc.gov/html/tlc/html/about/trip_record_data.shtml}} since this is the densest one and has 14 million trips. To obtain a reasonable geographical region representation, we utlize the taxizone shapefile again to segment Manhattan borough into 67 subregions, as is shown in Figure~\ref{fig:nyc_polygon}. Next, we select the length of time interval to be 15 minutes, i.e., 96 time intervals per day. Therefore, the NYC OD stochastic speed can be represented as an $\sR^{67\times 67 \times 7}$ tensor and 96 tensors per day.

\noindent
\textbf{Chengdu Data Set~(CD): }CD contains 1.4 billion GPS records from 14,864 taxis collected from 2014-08-03 to 2014-08-30 in Chengdu, China\footnote{\url{https://goo.gl/3VsEym}}. Each GPS record consists of a taxi ID, a latitude, a longitude, an indicator of whether the taxi is occupied, and a timestamp. We consider sequences of GPS records where taxis were occupied as trips. %as the parts occupancy status continuously to be ``1'' and 
%This gives 11,328,914 trips. 
%
%Next, we select the region within second ring road as our RoIs which has 3,636,845 trips, and 
We use a total of 3,636,845 trips that occurred within the second ring road of Chengdu.  
Next, we partition Chengdu within the second ring road into 79 regions according to the main roads; see Figure~\ref{fig:cd_polygon}. 
%We keep the same time interval length, 15 minutes, as previous dataset. 
The OD stochastic speeds for CD are represented as an $\sR^{79\times 79 \times 7}$ tensor. % and 96 tensors per day.

Table~\ref{tbl:ds} shows the statistics of the two datasets. Figures~\ref{fig:cd_cdf} and~\ref{fig:cd_cdf} show the speed distributions for both datasets. We use 70\% of the data for training, 10\% for validation, and the remaining 20\% for testing for both NYC and CD. 

%+++Training vs testing data sets: +++
%
%+++ Check the new baseline terms. NH, BF, AF, and FC. 

%\subsubsection{Model Functionalities}
\subsubsection{Forecast Settings}
%The proposed framework is a generic and end-to-end deep learning model of which the modules are changable with either classic neural network, e.g., FC, or state-of-the-art method, e.g., GCNN. 
%
%To fullfil the functionality of OD stochastic speed forecasting, we employ the idea of sequence to sequence model where
%
We consider settings where $s=3$ or $s=6$ while varying $h$ among 1, 2, and 3.
This means that we use stochastic OD matrices from 3 or 6 historical intervals to predict stochastic OD matrices during up to 3 future intervals, respectively. %This means that we fix $s=3$ or $s=6$, while varying $h$ among 1, 2, and 3. 
%
%select 6 previous time intervals as observations and 3 followed time intervals as forecasts in the experiments, i.e., $s=6$ and $h=3$. 
An example for $s=6$ and $h=3$ can be: given stochastic OD matrices in intervals [8:00, 8:15),  [8:15, 8:30),  [8:30, 8:45),  [8:45, 9:00),  [9:00, 9:15), and [9:15, 9:30), we predict stochastic OD matrices in intervals [9:30, 9:45),  [9:45, 10:00), and [10:00, 10:15). % when $s=6, h=3$.  % will be the results returned from our framework. 

%+++ Check the new baseline terms. NH, BF, AF, and FC. ++++

\subsubsection{Baselines}
%We compar FODS and GODS with 
To evaluate the effectiveness of the proposed base framework (BF) and the advanced framework (AF), we consider five baselines. 
(1) Naive Histograms~(NH): for each OD pair, we use all travel speed records for the OD pair in the training data set to construct a histogram and use the histogram for predicting the future stochastic speeds. 
Next, we model the stochastic speeds for each OD pair as a time series of vectors, where each vector represents the stochastic speed of the OD pair in an interval. Based on this time series modeling, we consider three time series forecasting methods:   
(2)~Support Vector Regression~(SVR)~\cite{basak2007support}, (3)~Vector Autoregression~(VAR)~\cite{sims1980macroeconomics}, and (4)~Gaussian Process Regression~(GP)~\cite{rasmussen2004gaussian}. 
%(2)~Support Vector Regression~(SVR): a regression method that maintains all the main features that characterize the algorithm. 
%
%(3)~Vector Autoregression~(VAR):  % where at each time stamp dimension corresponds to a 
%
%a stochastic process model that used to capture the linear interdependencies among multiple time series. In our setting, it tries to figure out the dependencies among different OD pairs.  (4)~Gaussian Process Regression~(GP): is a stochastic process whose realizations consist of random values associated with every point in a range of times~\cite{rasmussen2004gaussian}. 
(5)~Fully Connected~(FC): this is a variant of BF where we only directly use a fully connected layer to obtain a single dense tensor (instead of performing factorization into two dense tensors) to replace the factorization step in BF. 

%as is desribed in the previous section, FC here serves a role of encoding the raw signal into GRU and decoding the output of GRU to forecasts.

\subsubsection{Evaluation Metrics}
To quantify the effectiveness of the proposed frameworks, we use three commonly used distance functions that work for distributions, i.e., Kullback-Leibler divergence~(KL), Jensen-Shannon divergence~(JS), and earth-mover's distance~(EMD), to measure the accuracy of forecasts. %In the rest part, we report those measurements for each forecast step, separately.

Specifically, the general dissimilarity metric is defined as follows. 
%\begin{equation}
%\text{DisSim}_{\text{metric}}^{(k)} = \frac{\sum_{t=1}^{T}\text{weighted}_{\text{metric}}(\Omega^{(t+k)}, \tM^{(t+k)}, \widehat{\tM}^{(t+k)})}{\sum_{t=1}^{T}\Omega^{(t+k)}} 
%\end{equation}
%
%\begin{equation}
%\text{weighted}_{\text{metric}}(\Omega, \tM, \widehat{\tM}) =  \frac{\sum_{i=1}^{N}\sum_{j=1}^{N}\Omega_{i,j}\text{metric}(\tM_{i,j,:}, \widehat{\tM}_{i,j,:})}{\sum_{i=1}^{N}\sum_{j=1}^{N}\Omega_{i, j}}
%\end{equation}

\begin{equation}
\text{DisSim}_{\text{metric}}^{(k)} = \frac{\sum_{t=1}^{T}\sum_{i=1}^{N}\sum_{j=1}^{N'}\Omega_{i,j}^{(t+k)}\text{metric}(\tM_{i,j,:}^{(t+k)}, \widehat{\tM}_{i,j,:}^{(t+k)})}{\sum_{t=1}^{T}\sum_{i=1}^{N}\sum_{j=1}^{N}\Omega_{i, j}^{(t+k)}} , 
\end{equation}

%\begin{equation}
%\text{weighted}_{\text{metric}}(\Omega, \tM, \widehat{\tM}) =  \frac{\sum_{i=1}^{N}\sum_{j=1}^{N}\Omega_{i,j}\text{metric}(\tM_{i,j,:}, \widehat{\tM}_{i,j,:})}{\sum_{i=1}^{N}\sum_{j=1}^{N'}\Omega_{i, j}}
%\end{equation}
where $1 \le k \le h$ denotes the $k$-th step ahead forecasts, $T$ is the size of testing set. Next, $\Omega^{(t+k)}$, $\tM^{(t+k)}$, and $\widehat{\tM}^{(t+k)}$ are the indication matrix, ground truth tensor, and forcast tensor, respectively. $\Omega^{(t+k)}_{(i,j)}=1$ if observations exist from region $i$ to region $j$ at step $(t+k)$; otherwise, $\Omega^{(t+k)}_{(i,j)}=0$. Moreover, $\text{metric}(\cdot)$ is a generic metric function that can be any of the metrics mentioned above and defined next. 
For simplicity, we use $\vm$ and $\widehat{\vm} \in \sR^{K}$ to denote $\tM_{i,j,:}^{(t+k)}$ and $\widehat{\tM}_{i,j,:}^{(t+k)}$, respectively.  

KL divergence,
\begin{equation}
\text{KL}(\vm, \widehat{\vm}) = \sum_{k=1}^{K}\widehat{\vm}_k \text{log}(\frac{\widehat{\vm}_k + \delta}{\vm_k + \delta}),
\end{equation}
where $\delta$ is a positive small value to prevent having a zero when using the $log$ function. We use $\delta=0.001$ in the experiment. 

Jensen-Shannon divergence,
\begin{equation}
\text{JS}(\vm, \widehat{\vm}) = \frac{\text{KL}(\vm, \widehat{\vm}) + \text{KL}(\widehat{\vm}, \vm)}{2}
\end{equation}

Earth mover's distance,
\begin{equation}
\text{EMD}(\vm, \widehat{\vm}) = \frac{\sum_{i=1}^{K}\sum_{j=1}^{K}\mF_{i, j}d_{i, j}}{\sum_{i=1}^{K}\sum_{j=1}^{K}\mF_{i, j}},
\end{equation}
where flow matrix $\mF$ is the optimal flow that minimizes the overal cost from $\vm$ to $\widehat{\vm}$~\cite{rubner1998metric}.

All three functions capture the dissimilarity between an estimated and a ground-truth distribution. Thus, low values are preferred. % metrics describe the dissimilarities or distances of the estimated distribution from the ground truth distribuiton, and thus lower values are prefered. % and the lowest value for those metrics in each row are highlighted in \textbf{bold} font. 

\subsubsection{Model Construction}

%\subsubsection{Training}
The proposed frameworks are trained by minimizing the two loss functions defined in Equation~\ref{eq:loss_func_BF} for BF and Equation~\ref{eq:loss_func} for AF, using backpropagation. We use the Adam optimizer due to its good performance. The hyper-parameters were configured manually based on the loss on a separate validation set. Specifically, we set the initialization learning rate to 0.001, set with the decay rate to 0.8 at every 5 epochs, and set the dropout rate to 0.2. 

Table~\ref{tbl:model_construct} shows the optimal configurations of the hyper-parameters for the three deep learning methods and numbers of weight parameters used in each model for both datasets. 
Baseline FC first encodes the input into a 2D latent space via an FC operation, denoted as ${FC}_{2}$. Then it calls a GRU with 3 units and 1 layer to capture the temporal dynamics, denoted as ${GRU}_{3}^{1}$. Finally, another FC is called to project the output from GRU to an OD stochastic tensor with the following dimensions: $\#\text{Source Regions} \times \#\text{Destination Regions} \times \#\text{Buckets}$, e.g., $67\times 67 \times 7=31,423$ dimensions for NYC, denoted as ${FC}_{31,423}$. 
For BF and AF, we apply two identical configurations for origin and destination factorization, which is why we have ``$2\times$'' on the first configuration, respectively. 
For BF, we first utilize ${FC}_2$ to encode the input for the first factorization. Then we adopt ${GRU}_{2}^{1}$ to learn the temporal dynamics. At the end of GRU, we project the output into a corresponding factorization with the following dimensions: $\#\text{Source Regions} \times r \times \#\text{Buckets}$, where $r$ is the rank of the factorized dense matrix which we set to 5, e.g., $67 \times 5 \times 7=2,345$ for NYC, denoted as ${FC}_{2,345}$. 
The configuration for AF is very different from the previous two models. First, we adopt two combinations of GCNN, ${GC}_{K}^{Q}$, where $Q$ is the filter number and $K$ is the filter size, and pooling operation, $P{p}$, where $p$ is the pooling size, e.g., ${GC}_{8}^{32}$-$P4$-${GC}_{4}^{32}$-$P2$ for NYC. Then, the encoded features are fed into a CNRNN with $n$ layers where each layer has four Cheby-Nets. Assuming that the number and size of the filters are $Q_c$ and $K_c$,  this operation can be written as ${GCR}_{n}^{Q_c\times K_c}$, e.g., ${GCR}_{2}^{32 \times 4}$, implying 2 CNRNNs where the GCNN in each gate has 32 graph convolutional filters of size 4.

From the above configurations, although AF uses the most complex models, AF uses the fewest weight parameters (see the \#  weights column in Table~\ref{tbl:model_construct}).

\begin{table}
	\centering
	\normalsize
	\begin{tabular}{ c|c | cc}
		\hline
		Data & Model & Configuration &\#Weights \\
		\hline\hline
		\multirow{3}{2em}{NYC}
		& FC & ${FC}_{3}$-${GRU}_{3}^{1}$-${FC}_{31,423}$ & 408,535 \\
		& BF & $2\times$${FC}_{2}$-${GRU}_{2}^{1}$-${FC}_{2,345}$& 391,182    \\
		& BF & 2 $\times$ ${GC}_{8}^{32}$-$P4$-${GC}_{4}^{32}$-$P2$-${GCR}_{2}^{32\times4}$ &  339,726  \\\hline
		\multirow{3}{2em}{CD}
		& FC & ${FC}_{3}$-${GRU}_{3}^{1}$-${FC}_{43687}$ & 567,967  \\
		& BF &  $2\times$${FC}_{2}$-${GRU}_{2}^{1}$-${FC}_{2,765}$ & 415,339  \\
		& AF & 2 $\times$ ${GC}_{8}^{32}$-$P4$-${GC}_{4}^{32}$-$P2$-${GCR}_{2}^{32\times4}$ & 367,502  \\ \hline
	\end{tabular}
	\caption{Model Construction and Hyper-Parameter Selection}
	\label{tbl:model_construct}
	\vspace{-15pt}
\end{table}

\subsection{Experimental Results}
%We proceed to conduct experiments to verify the effectiveness of our proposed method against the baseline methods using NYC and CD datasets. We compare our method with baseline methods and report on the above mentioned threes metrics, KL, JS and EMD, in multi-step ahead, time varying, and distance varying forecasting. Moreover, we take a look into of how $k$ selection influence on GODS. 
\subsubsection{Overall Results}
%We vary $h$, i.e., $h$-interval ahead forecasting, to study the effect of $h$. 
%
%Table~\ref{tbl:whole_results} shows the comparison of evaluated methods for OD stochastic speed forecasting using NYC and CD datasets with varying horizons, i.e., 0-15 mins, 15-30 mins, and 30-45 mins, respectively, and different evaluatation metrics, i.e., KL, JS, EMD, respectively. 
%
We compare the accuracies of the different methods, using KL, JS, and EMD to evaluate the forecast accuracy; see in Tables~\ref{tbl:whole_results_s3} and~\ref{tbl:whole_results_s6}. We also vary $s$, i.e., the number of historical stochastic speed matrices, and $h$, i.e., the $h$-intervals ahead forecasting, to study the effect of $s$ and $h$.   
We have the following observations. 
(1) The deep learning based methods perform better than the other baselines in most cases. %This indicates that deep learning model has some superior performance, but it calls for theorical support to design. 
%(2) The proposed methods, BF an AF, outperform other baselines for all two settings of $s$, all three settings of $h$, and all three evaluation metrics. This indicates that the proposed frameworks, which involve factorization and RNN based forecasting, are effective for OD matrix forecasting in settings with data sparseness.
(2) The proposed basic framework BF performs better than other methods in most settings. % but only fails once when $s=3$ and $h=3$ for KL metric. 
This indicates that the proposed frameworks, which involve factorization and RNN based forecasting, are effective for OD matrix forecasting in settings with data sparseness. 
(3) The advanced framework AF is significantly better than other methods, including BF, in all settings. This suggests that by taking into account the spatial correlations among regions using two GCNNs, the learned features become more meaningful, which then improves forecasting accuracy. 
%(4) The reuslts of our methods on different horizons do not show a significant difference in both datasets over three metrics. Considering the results on HA method seem invariant in all three horizons on both datasets, the reason lies in the stochastic speed from region to region within 45 minutes does not vary a lot, i.e., the temporal dynamics is quite gentle. Thus, the temporal states are well transfered from one to next time step. 
(4) The reuslts on NYC are better than those on CD. This is because the regions in NYC are more homogeneous (i.e., within Manhatten)
than the regions in CD that cover a much larger and more diverse region. This in turn makes the traffic situations in CD much more complex and more challenging to forecast. 
(5) When varying $h$, the accuracy of AF becomes worse, i.e., larger metric values. This suggests that forecast far into the future becomes more challenge. 
(6) When fixing $h$, we compare the two tables and observe that the performance of AF is better at $s=3$ than $s=6$. This seems to indicate that the traffic variations are more dependent on short-term history (i.e., $s=3$) than on long-term history (i.e., $s=6$). 
 % This seems to suggest that having data in six historical intervals is sufficient to predict the stochastic speeds in the future 45 mins.
%(5) When varying $h$, the accuracy does not vary much for any method. This seems to suggest that having data in six historical intervals is sufficient to predict the stochastic speeds in the future 45 mins.
 %, the traffic seems %The reuslts of our methods on different horizons do not show a significant difference in both datasets over three metrics. Considering the results on HA method seem invariant in all three horizons on both datasets, the reason lies in the stochastic speed from region to region within 45 minutes does not vary a lot, i.e., the temporal dynamics is quite gentle. Thus, the temporal states are well transfered from one to next time step. 

According to the above results, in the following, we only consider FC, BF, and AF, and we only consider the setting where $h=1$ and $s=6$, i.e., 1-step ahead forecasting with 6 historical observations. % performance are significantly better than other methods. Also, we only show the results of one step forecasting, i.e. 0-15 minutes ahead, due to their similarity. 

\begin{table}[t]
\centering
\footnotesize
\begin{tabular}{ c||c|c | ccccccc}
\hline
Data& Metric  & $h$  & NH & SVR & VAR & GP & FC & BF & AF  \\
\hline\hline
\multirow{9}{1em}{NYC} 
& \multirow{3}{1em}{KL}
    & 1 	& 0.592 	& 0.704	& 0.554	 & 0.522	 & 0.446	 &  0.427	 & \textbf{0.311} \\
 & & 2 	&  0.592	& 0.713 & 0.562	 	& 0.535	 & 0.438 	 &  0.417	 &  \textbf{0.313}	\\
 & & 3 	& 0.592 	& 0.720	 & 0.577	 & 0.545 	 & 0.438	 &  0.415   	&   \textbf{0.314} \\ \cline{2-10}
& \multirow{3}{1em}{JS}
    & 1 	& 0.439	& 0.530	&  0.440	 & 0.391	 & 0.332 	 &  0.322	 & \textbf{0.299} \\
 & &2	& 0.439 	& 0.537	 & 0.452	 & 0.400	 & 0.327 	 &  0.317	 &  \textbf{0.302}	\\
 & & 3 	& 0.439 	&0.543 	 & 0.471	 & 0.408	 & 0.327	 &  0.316   & \textbf{0.305} \\\cline{2-10}
& \multirow{3}{1em}{EMD}
    & 1 	& 0.293 	 & 0.452	& 0.305	 & 0.266 	 & 0.250 	 &  0.246 	 &  \textbf{0.214}  \\
 & & 2 	& 0.293	 & 0.455	 & 0.313	 & 0.270 	 & 0.247 	 &  0.243	 &  \textbf{0.216}	\\
 & & 3 	& 0.293 	 & 0.458 	 & 0.317	 & 0.274	 & 0.247	 &  0.243      &   \textbf{0.217} \\ \hline \hline
 \multirow{9}{1em}{CD} 
 & \multirow{3}{1em}{KL}
    & 1 	& 0.697 	& 0.818 	& 0.699	 & 0.674	 & 0.694	 & 0.582	& \textbf{0.549} \\
 &    &2 	& 0.709 	& 0.836	 &  0.715 	&0.687 	 & 0.689 	 &  0.586	 & \textbf{0.555} \\
  &   & 3 & 0.700 		& 0.822	 & 0.780	 &0.684 	 & 0.807 	 &  0.792	 &  \textbf{0.626} \\ \cline{2-10}

& \multirow{3}{1em}{JS}
  & 1 	& 0.580 	& 0.672	& 0.814	 &0.597 	 & 0.517 	 & 0.438 	&   \textbf{0.435}\\
 & & 2 	& 0.584 	& 0.677	 & 0.869	 & 0.599	 & 0.513 	 &  0.442	 &  \textbf{0.444}  \\
 & & 3 & 0.592 	& 0.692	 & 0.875	 & 0.619	 & 0.590 	 &  0.571	 &  \textbf{0.500} \\ \cline{2-10}
& \multirow{3}{1em}{EMD}
  & 1 	& 0.441 	& 0.543	& 0.799	 &0.439 	 & 0.360	 & 0.307    & \textbf{0.289} \\ 
  & & 2 	& 0.443 	& 0.539	 & 0.871	 & 0.434 	 & 0.361	 &  0.313   & \textbf{0.295} \\ 
 & &3 & 0.464 	& 0.574	 & 0.787	 & 0.471 	 & 0.423	 &  0.378   & \textbf{0.311} \\ \hline
\end{tabular}
\caption{Forecast Accuracy with Varying $h$,  $s=3$.}
\label{tbl:whole_results_s3}
\vspace{-20pt}
\end{table}

\begin{table}[t]
\centering
\footnotesize
\begin{tabular}{ c||c|c | ccccccc}
\hline
Data & Metric & $h$ & NH & SVR & VAR & GP & FC & BF & AF  \\
\hline\hline
\multirow{9}{1em}{NYC} 
& \multirow{3}{1em}{KL}
    & 1 & 0.592 	& 0.698	& 0.566	 & 0.522	 & 0.453 	 &  0.437	 & \textbf{0.343} \\
 & & 2 &  0.592	& 0.706	 & 0.576	 & 0.535	 & 0.445 	 &  0.421	 &  \textbf{0.347}	\\
 & & 3 & 0.593 	& 0.713	 & 0.587	 & 0.545 	 & 0.440	 &  0.410   	&   \textbf{0.348} \\ \cline{2-10}
& \multirow{3}{1em}{JS}
    & 1 & 0.440	& 0.524	&  0.442	 & 0.394	 & 0.337 	 &  0.327	 & \textbf{0.305} \\
 & & 2 & 0.440 	& 0.531	 & 0.455	 & 0.404	 & 0.333 	 &  0.318	 &  \textbf{0.308}	\\
 & & 3 & 0.440 	&0.537 	 & 0.472	 & 0.411 	 & 0.331	 &  0.313   & \textbf{0.310} \\\cline{2-10}
& \multirow{3}{1em}{EMD}
    & 1 & 0.293 	& 0.447	& 0.299	 & 0.265 	 & 0.248 	 &  0.238 	 &  \textbf{0.214}  \\
 & &2 & 0.293 	& 0.450	 & 0.305	 & 0.270 	 & 0.246	 &  0.235	 &  \textbf{0.216}	\\
 & & 3 & 0.294 & 0.452 	 & 0.314	 & 0.274 	 & 0.245	 &  0.233      &   \textbf{0.217} \\ \hline \hline
 \multirow{9}{1em}{CD} 
 & \multirow{3}{1em}{KL}
    & 1 & 0.686 	& 0.804 	& 0.741	 & 0.671	 & 0.753 	 & 0.668	& \textbf{0.600} \\
 & & 2 & 0.685 	&  0.806	& 0.759	 &0.671 	 & 0.740	 & 0.666 	&   \textbf{0.605}\\
 & & 3& 0.686 	&0.807	& 0.739	 &0.674  	 & 0.735	 & 0.667    & \textbf{0.609} \\ \cline{2-10}
& \multirow{3}{1em}{JS}
    & 1 & 0.578 	& 0.674	&  0.808 	 &0.610 	 & 0.554 	 &  0.517	 & \textbf{0.466} \\
 & & 2 & 0.577 	& 0.676	 & 0.850	 &0.610	 & 0.546 	 &  0.516	 &  \textbf{0.475}  \\
 & & 3 & 0.578 	& 0.678	 & 0.903	 &0.612 	 & 0.543	 &  0.517   & \textbf{0.483} \\ \cline{2-10}
& \multirow{3}{1em}{EMD}
    & 1& 0.449 	& 0.555	 & 0.716	 & 0.449 	 & 0.394 	 &  0.344	 &  \textbf{0.304} \\
 & & 2 & 0.448 	& 0.557	 & 0.767	 & 0.448	 & 0.389 	 &  0.344	 &  \textbf{0.305} \\
 & & 3 & 0.448 	& 0.559 	 & 0.890	 & 0.448	 & 0.386	 &  0.344   & \textbf{0.306} \\ \hline
\end{tabular}
\caption{Forecast Accuracy with Varying $h$,  $s=6$.}
\label{tbl:whole_results_s6}
\vspace{-20pt}
\end{table}

\subsubsection{Effect of Time of Day}
\begin{figure}[t]
	\centering
	\subfigure[NYC]{
		\begin{minipage}[b]{0.22\textwidth}
			\includegraphics[width=1\textwidth]{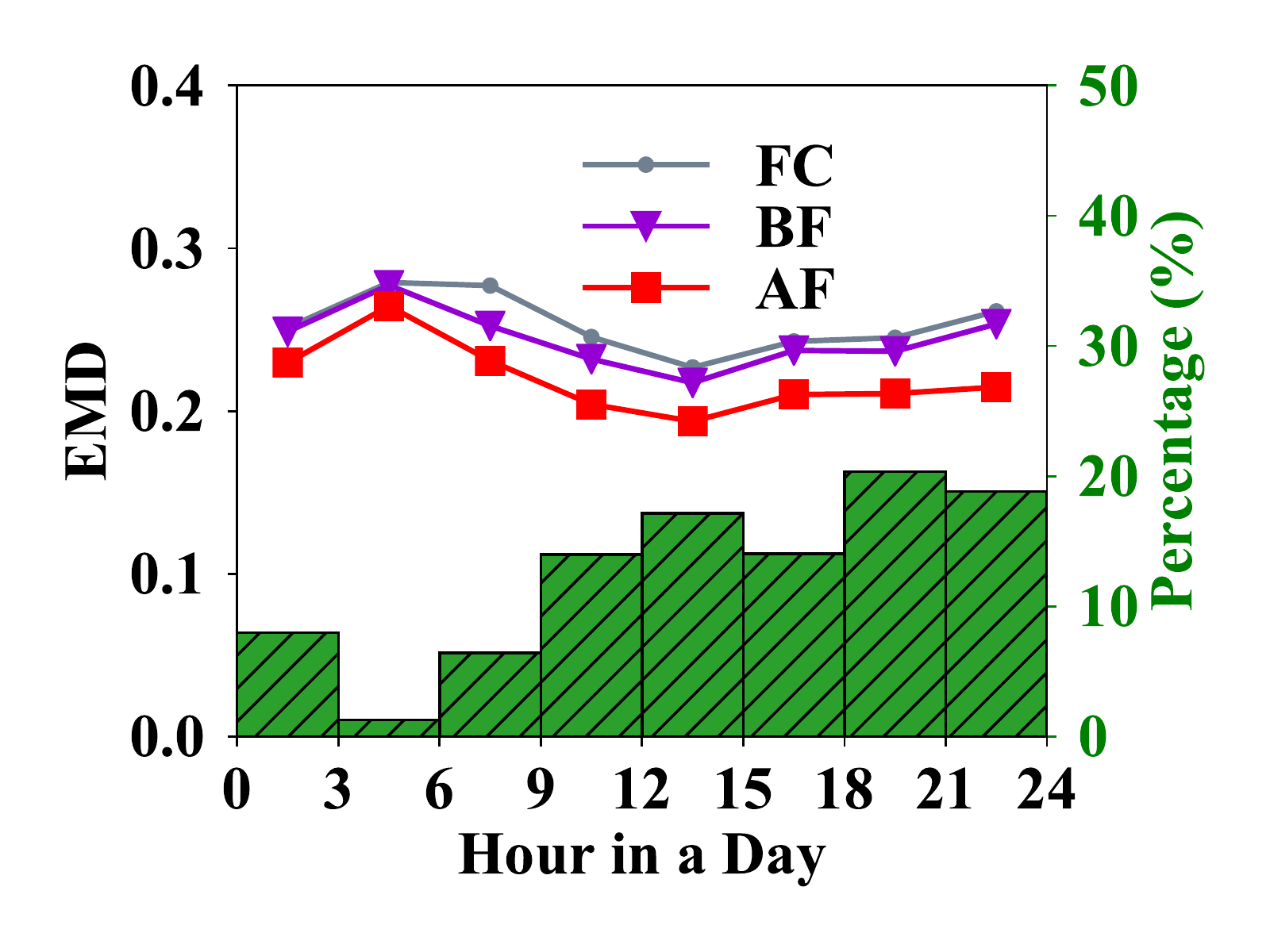}
		\end{minipage}
		\label{fig:nyc_time_emd}
	}
	\subfigure[CD]{
		\begin{minipage}[b]{0.22\textwidth}
			\includegraphics[width=1\textwidth]{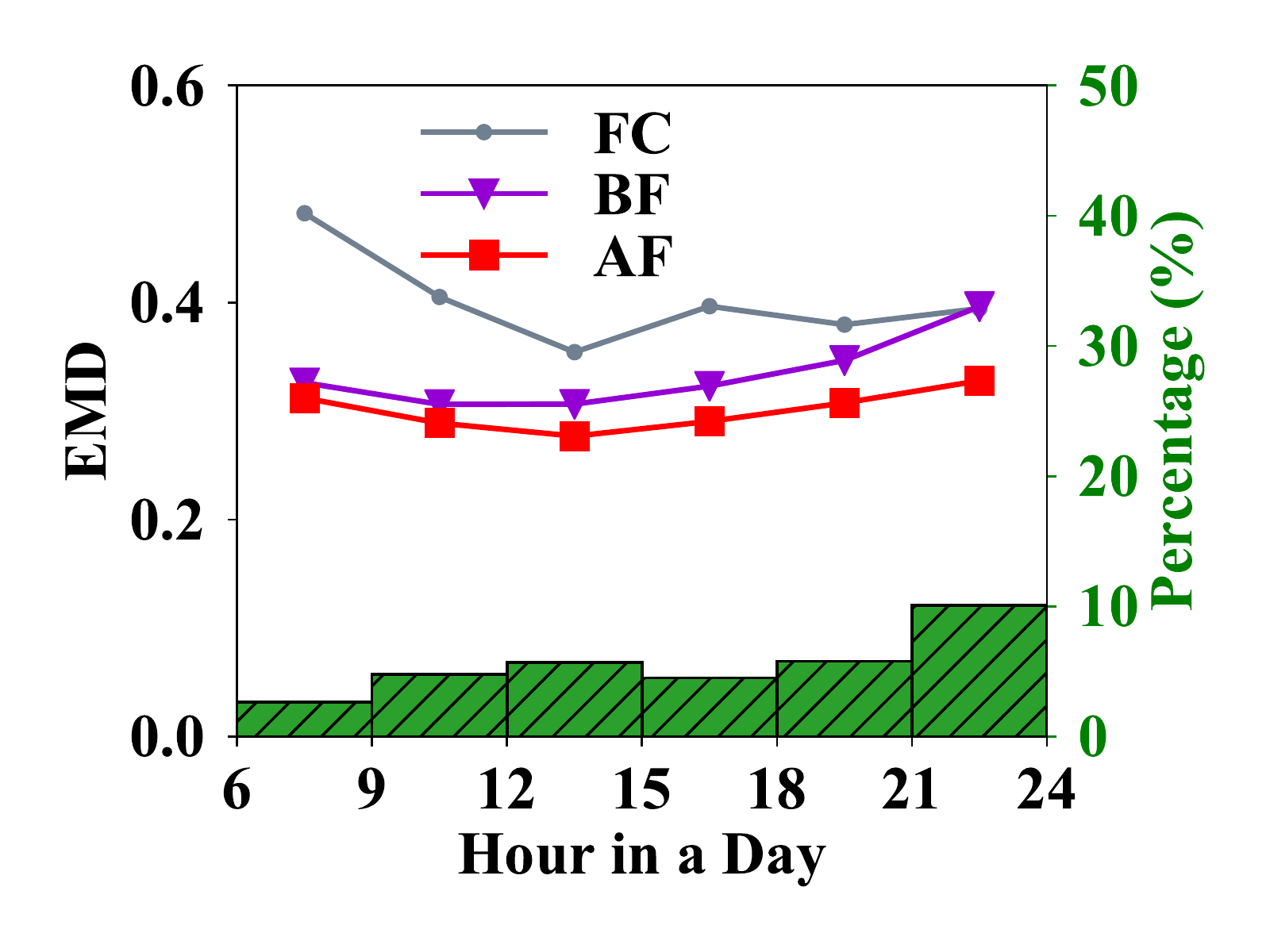}
		\end{minipage}
		\label{fig:cd_time_emd}
	}
	\caption{Effect of Time of a Day, EMD.}
	\label{fig:time_emd}
	\vspace{-10pt}
\end{figure}

\begin{figure}[t]
	\centering
	\subfigure[NYC]{
		\begin{minipage}[b]{0.22\textwidth}
			\includegraphics[width=1\textwidth]{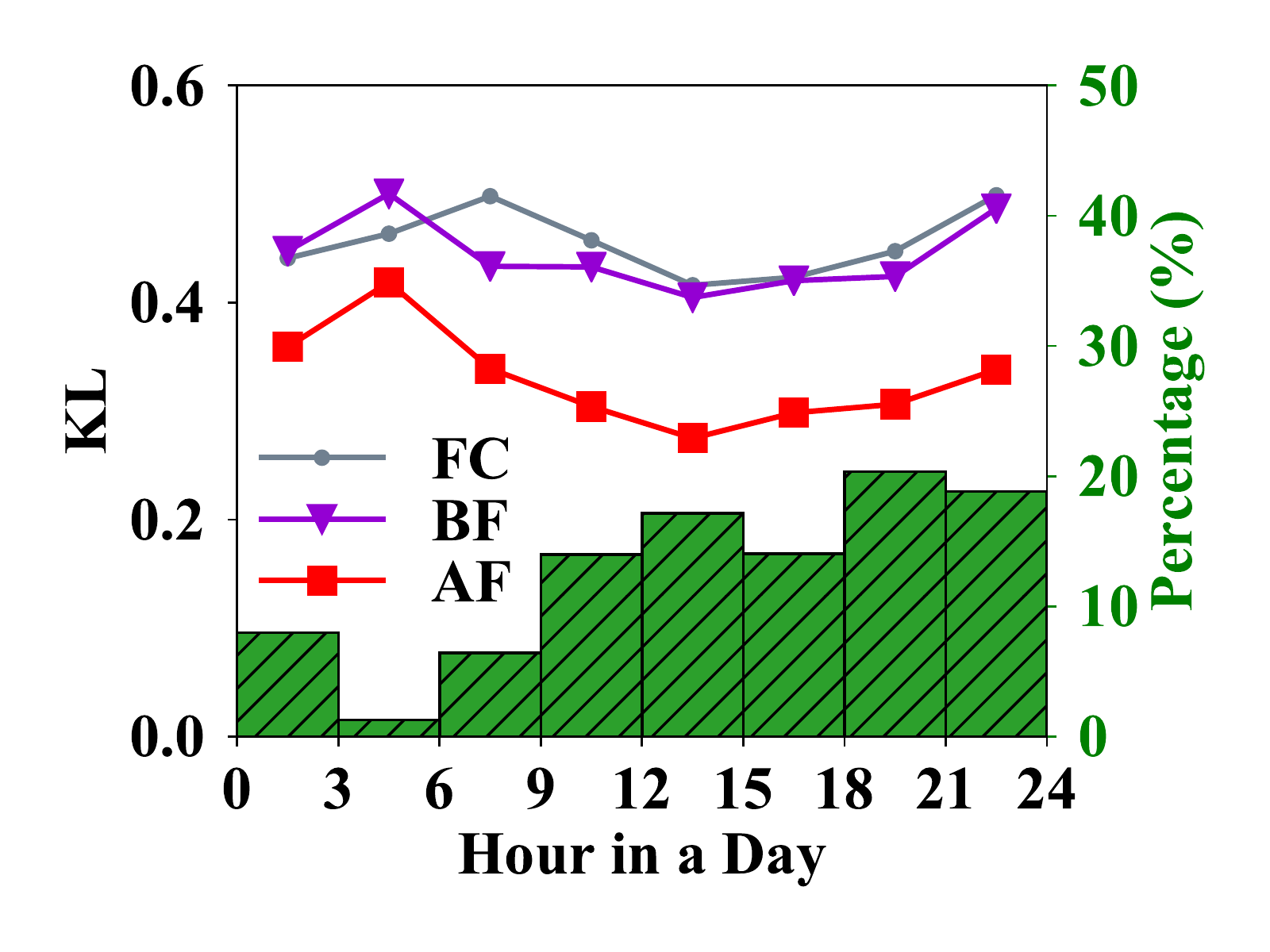}
		\end{minipage}
		\label{fig:nyc_time_kl}
	}
	\subfigure[CD]{
		\begin{minipage}[b]{0.22\textwidth}
			\includegraphics[width=1\textwidth]{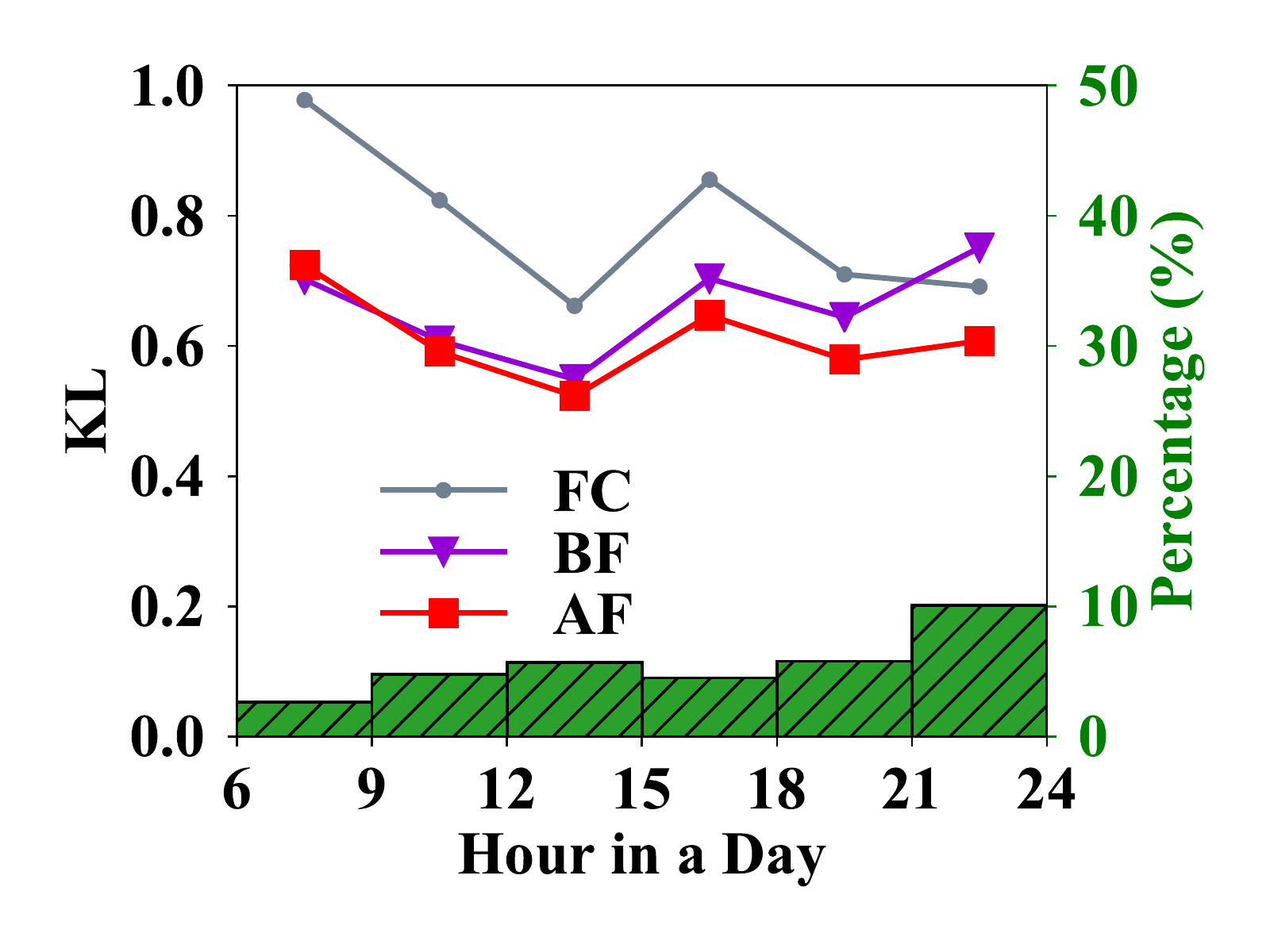}
		\end{minipage}
		\label{fig:cd_time_kl}
	}
	\caption{Effect of Time of a Day, KL.}
	\vspace{-10pt}
	\label{fig:time_kl}
\end{figure}

\begin{figure}[t]
	\centering
	\subfigure[NYC]{
		\begin{minipage}[b]{0.22\textwidth}
			\includegraphics[width=1\textwidth]{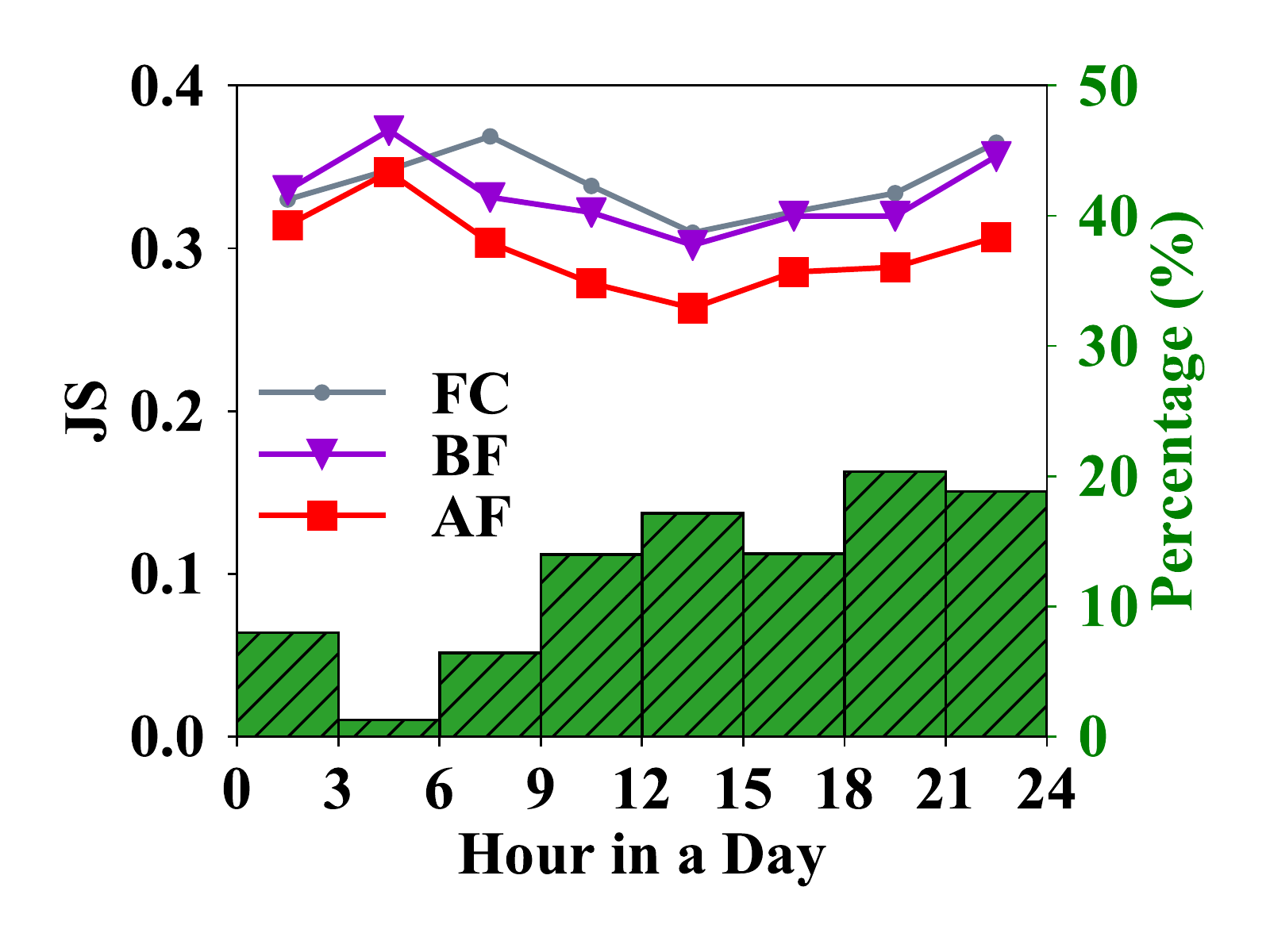}
		\end{minipage}
		\label{fig:nyc_time_js}
	}
	\subfigure[CD]{
		\begin{minipage}[b]{0.22\textwidth}
			\includegraphics[width=1\textwidth]{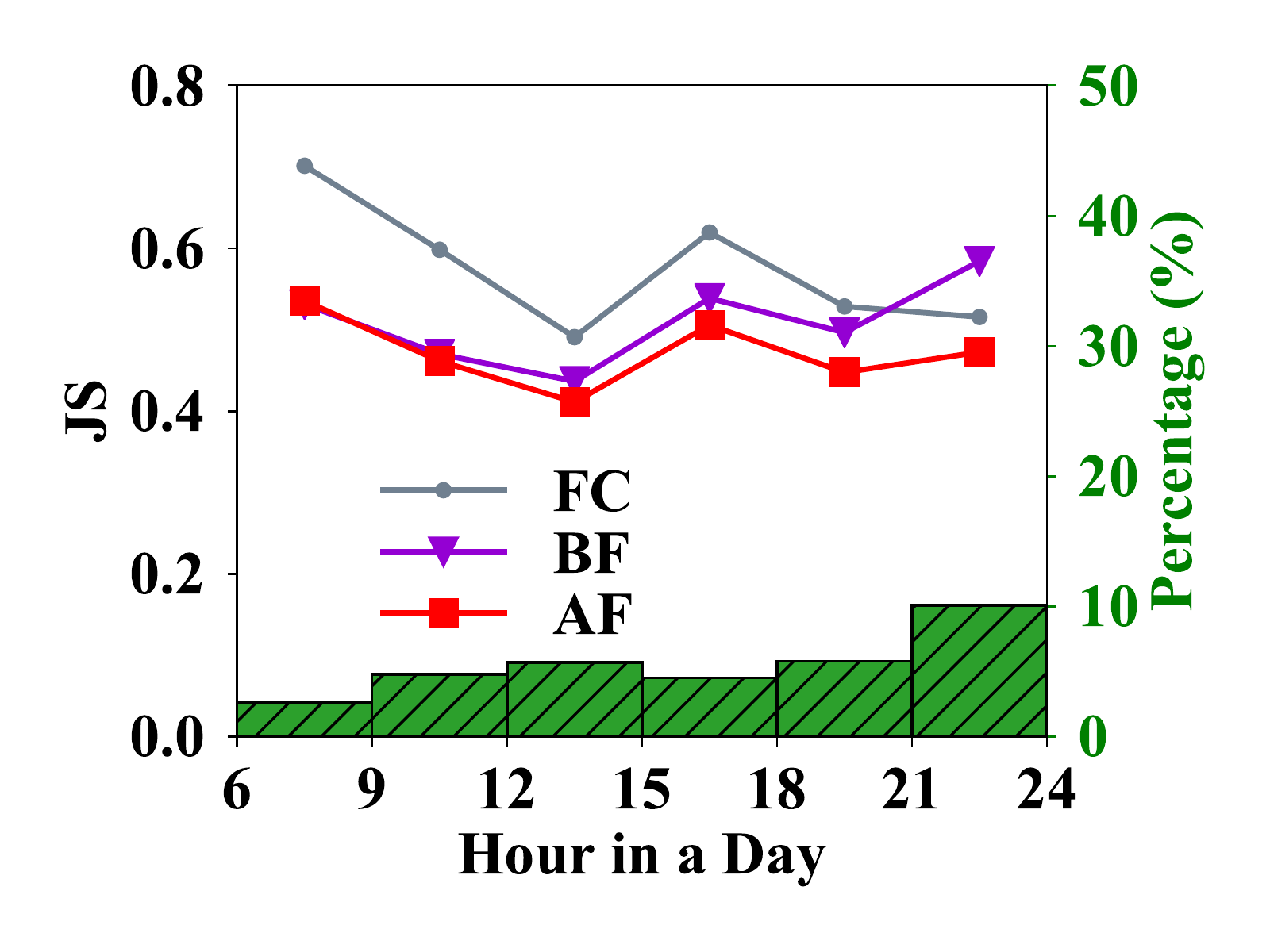}
		\end{minipage}
		\label{fig:cd_time_js}
	}
	\caption{Effect of Time of a Day, JS.}
	\label{fig:time_js}
	\vspace{-10pt}
\end{figure}

In this experiment, we aim at investigating forecasting performance for different intervals during a day. To this end, we show the forecast accuracy across different time intervals. % the time axis. 
Figures~\ref{fig:time_emd},~\ref{fig:time_kl}, and~\ref{fig:time_js} show the performance on both data sets when using EMD, KL, and JS. To visualize the results across time, we aggregate the results per each 3 hours. %This means that if the predicted OD matrix is for an interval within 0:00 to 3:00, the corresponding accuracy is aggregated into the  , e.g., forecasting results from 00:00 to 03:00 are averaged and is represented as the first data point in NYC results. 
We use three curves to represent the accuracy of FC, BF, and AF. In addition, we use bars to represent the percentages of data we have per each 3 hours. 
%
%dual-y axises to show both the metric results and data percentage at the same time, where the left y axis represents the metric value and the right y axis denotes the data percentage which is in green color and lies in the bottom of the figure. 
%
CD does not contain any data from 00:00 to 06:00, which is why the figures for CD start at 6. %so we only report the results from 07:00 to 24:00. 

Figures~\ref{fig:nyc_time_emd} and~\ref{fig:cd_time_emd} show the accuracy based on EMD. We observe that both AF and BF outperform FC in almost all the time intervals. This suggests the effectiveness of factorization in the proposed framework when contending with data sparseness. In addition, AF has the best performance and differs clearly from FC and BF. This suggests that by further capturing spatial and spatio-temporal correlations improves the forecast accuracy. 
%
%The advanced GODS, of capturing spatial correlation further improves its performance. 
%

We observe that the EMD for all the three methods is the worst in NYC during [3:00, 6:00). This is because the amount of testing data during [3:00, 6:00) is quite small, only accounting for around 1\% of the total testing data. 
%
%On both data sets, we observe that the curves go down during periods [6:00, 9:00) and [9:00, 12:00). This  which accounts for can be accounted for either increasing number of data or the peak hour increases the stochastic in speed. 
%
On both data sets, the best EMD values appear during [12:00, 15:00). This indicates that the traffic conditions during this time period seems to have the least dynamics thus making the forecasting less challenging. 
%
%Then, number of taxi trips starts to increase after 18:00 in both datasets since it's time to get off the work. Correspondingly, EMD starts to increase since the traffic system becomes more complicated. 
%
Similar trends can be observed when using KL and JS, as shown in Figures~\ref{fig:time_kl} and~\ref{fig:time_js}. Overall, the advanced framework AF achieves consistently the best forecasting performance on both datasets and on the three different evaluation metrics. More data enables more accurate forecasting. 

\subsubsection{Effect of Distances}
\begin{figure}[t]
	\centering
	\subfigure[NYC]{
		\begin{minipage}[b]{0.22\textwidth}
			\includegraphics[width=1\textwidth]{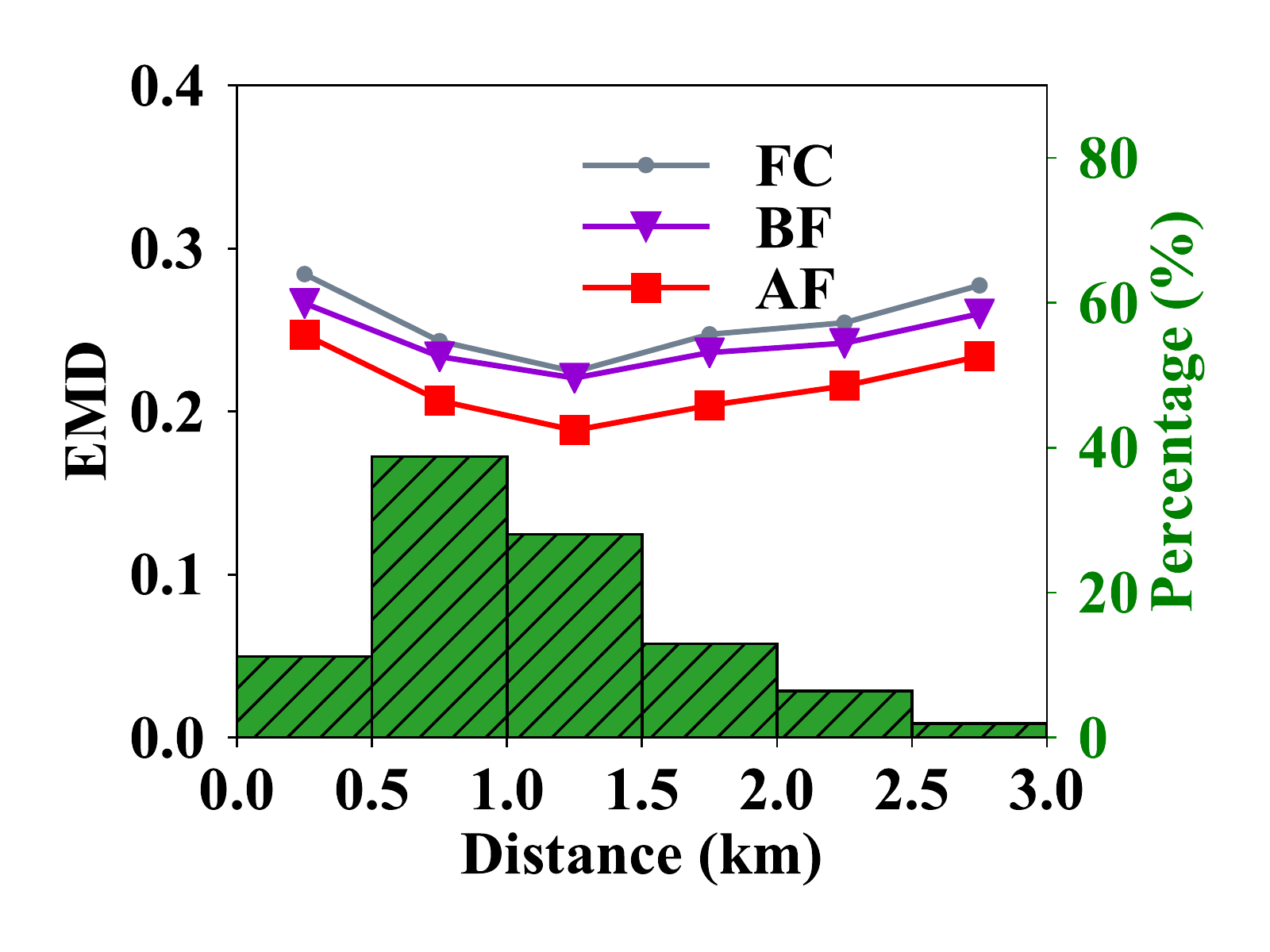}
		\end{minipage}
		\label{fig:dist_emd_nyc}
	}
	\subfigure[CD]{
		\begin{minipage}[b]{0.22\textwidth}
			\includegraphics[width=1\textwidth]{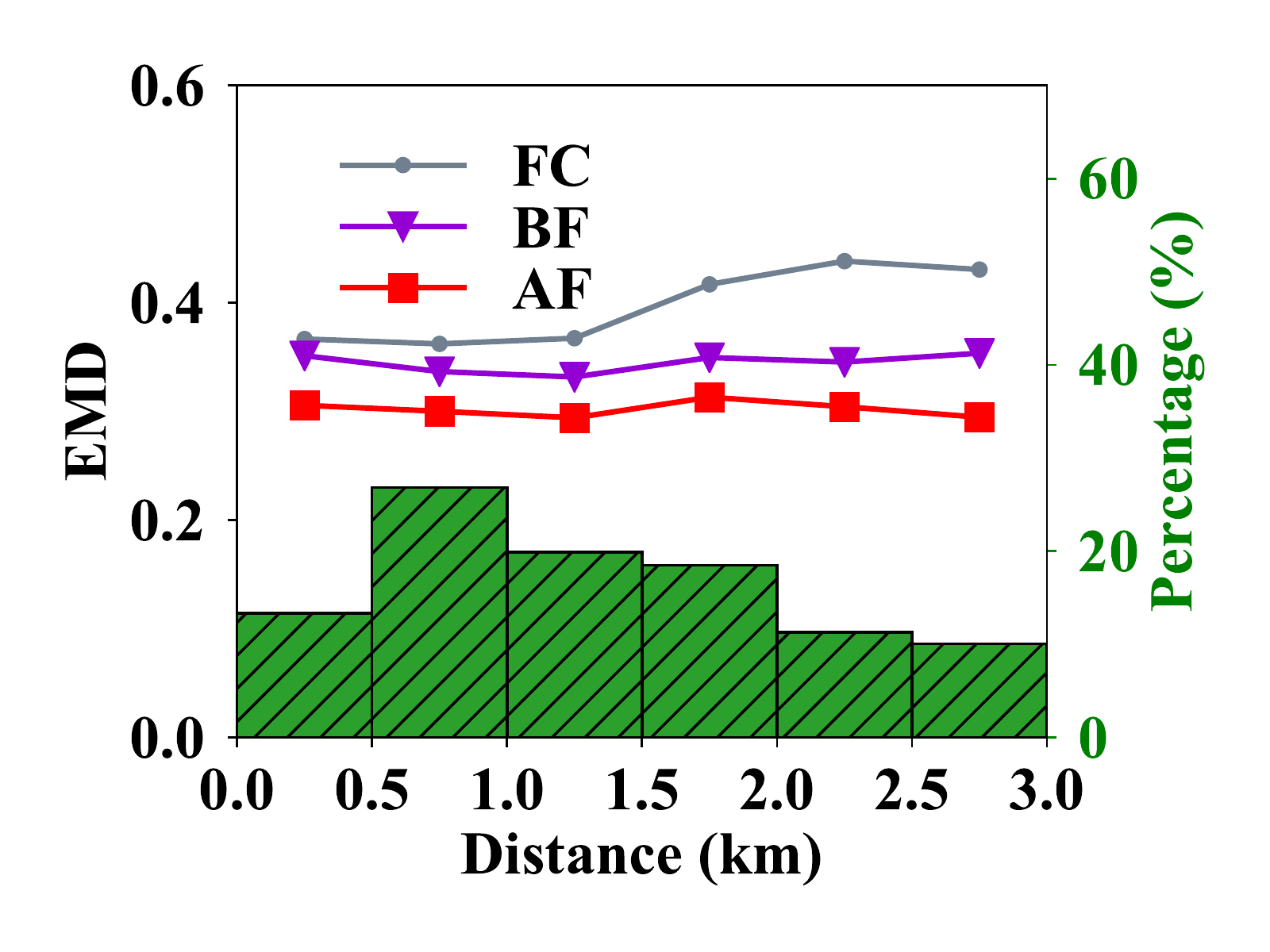}
		\end{minipage}
		\label{fig:dist_emd_cd}
	}
	\caption{Effect of Distances, EMD.}
	\label{fig:dist_emd}
	\vspace{-10pt}
\end{figure}

\begin{figure}[t]
	\centering
	\subfigure[NYC]{
		\begin{minipage}[b]{0.22\textwidth}
			\includegraphics[width=1\textwidth]{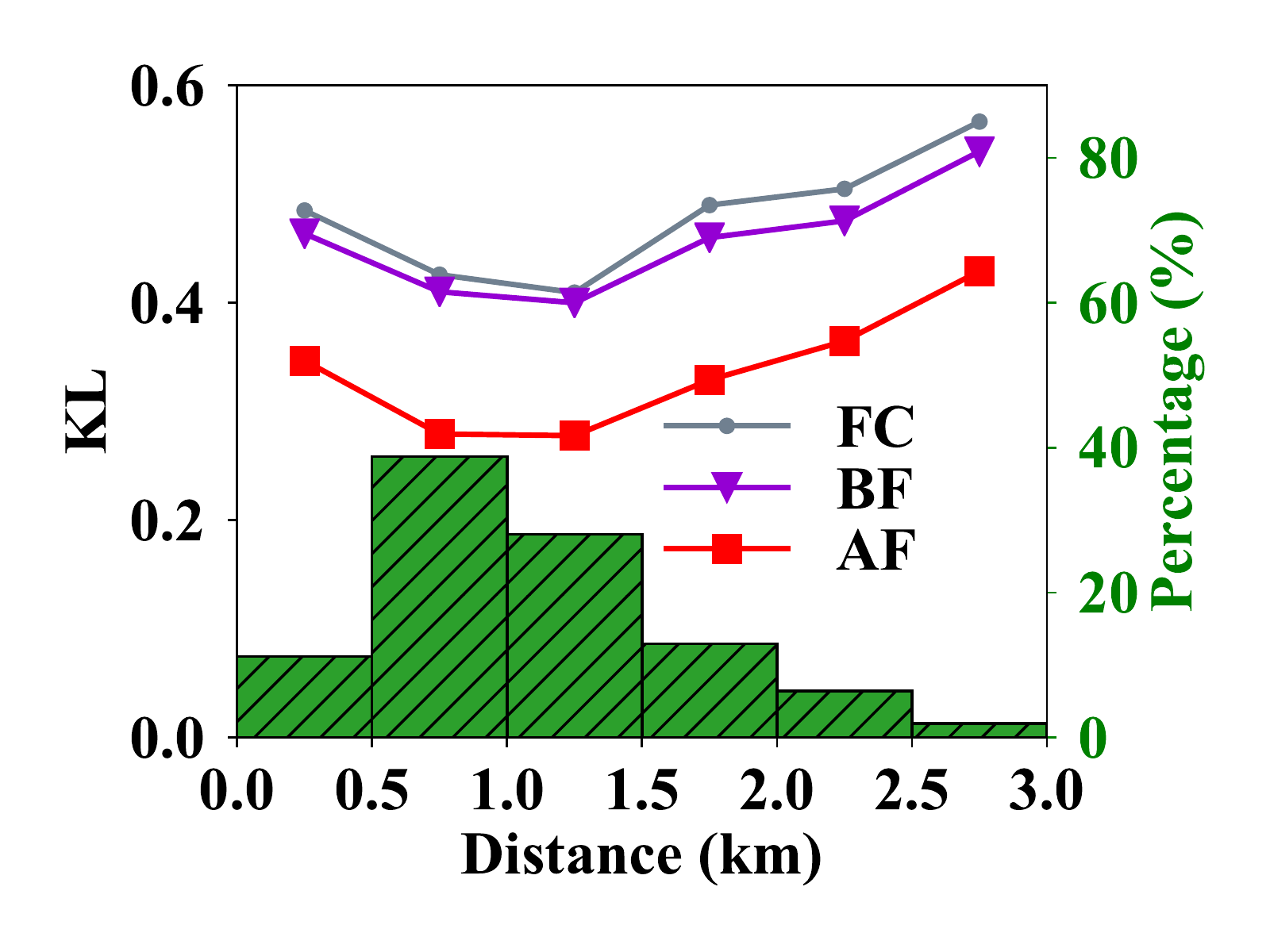}
		\end{minipage}
		\label{fig:dist_kl_nyc}
	}
	\subfigure[CD]{
		\begin{minipage}[b]{0.22\textwidth}
			\includegraphics[width=1\textwidth]{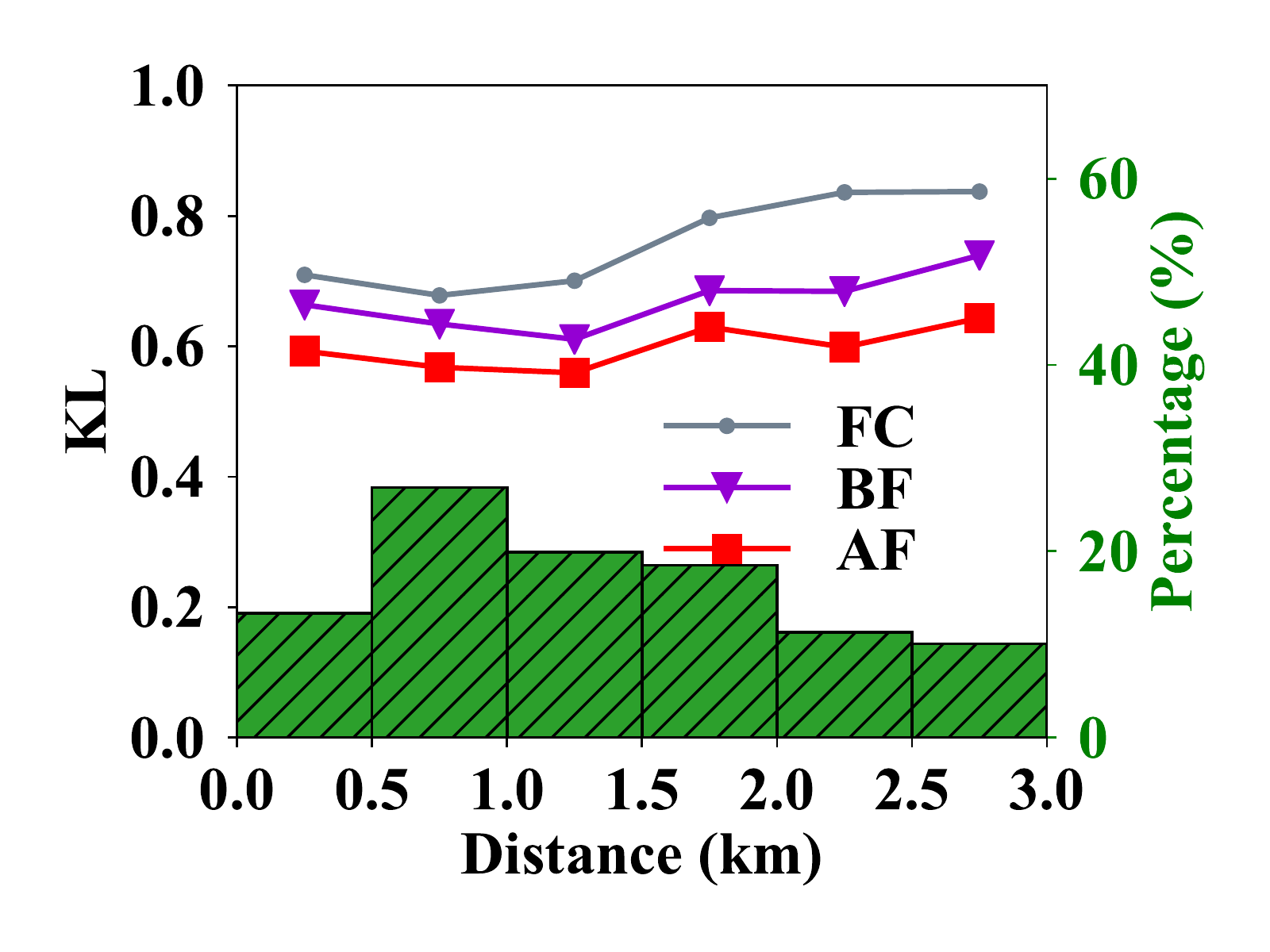}
		\end{minipage}
		\label{fig:dist_kl_cd}
	}
	\caption{Effect of Distances, KL.}
	\vspace{-10pt}
	\label{fig:dist_kl}
\end{figure}

\begin{figure}[t]
	\centering
	\subfigure[NYC]{
		\begin{minipage}[b]{0.22\textwidth}
			\includegraphics[width=1\textwidth]{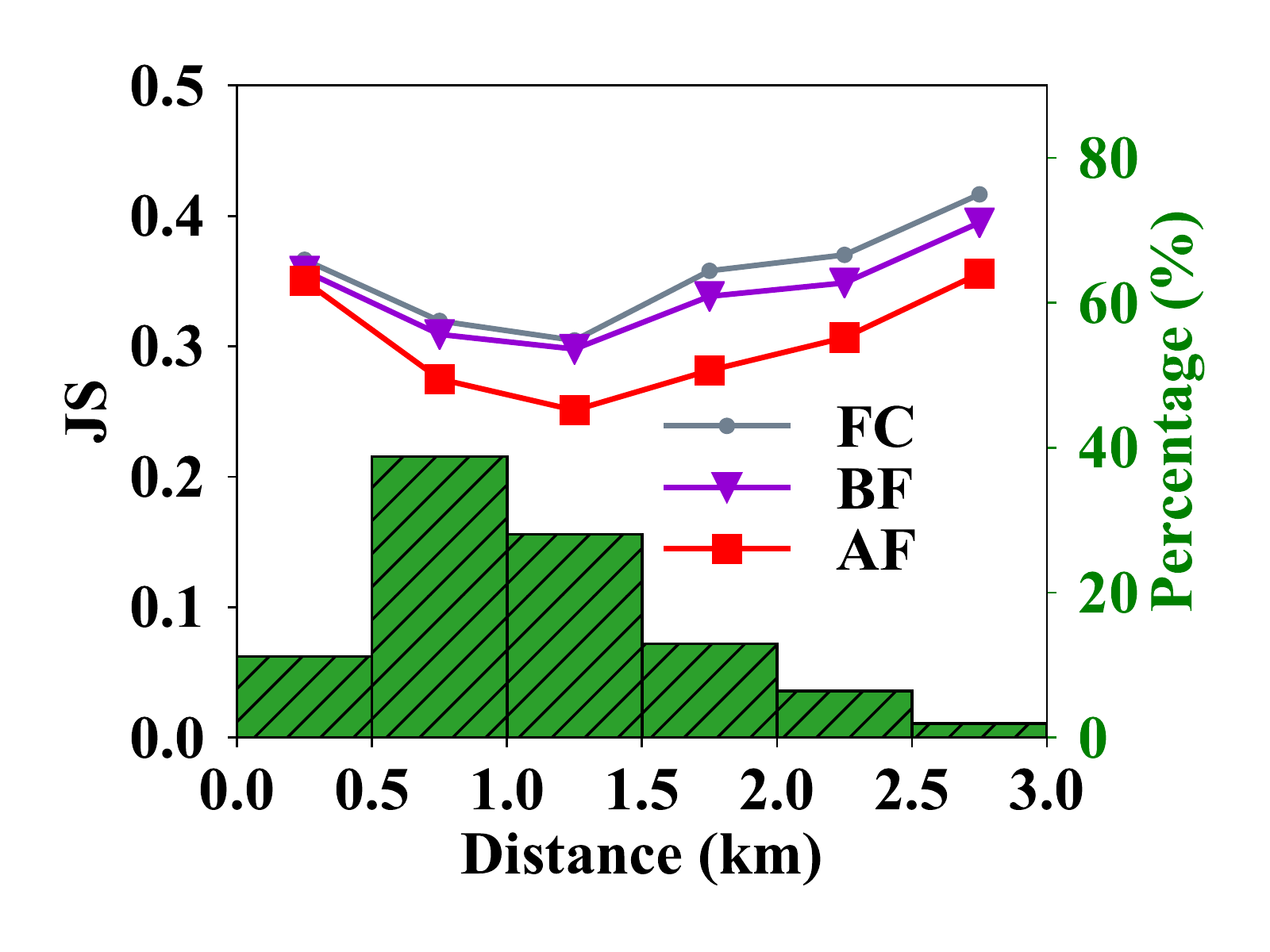}
		\end{minipage}
		\label{fig:dist_js_nyc}
	}
	\subfigure[CD]{
		\begin{minipage}[b]{0.22\textwidth}
			\includegraphics[width=1\textwidth]{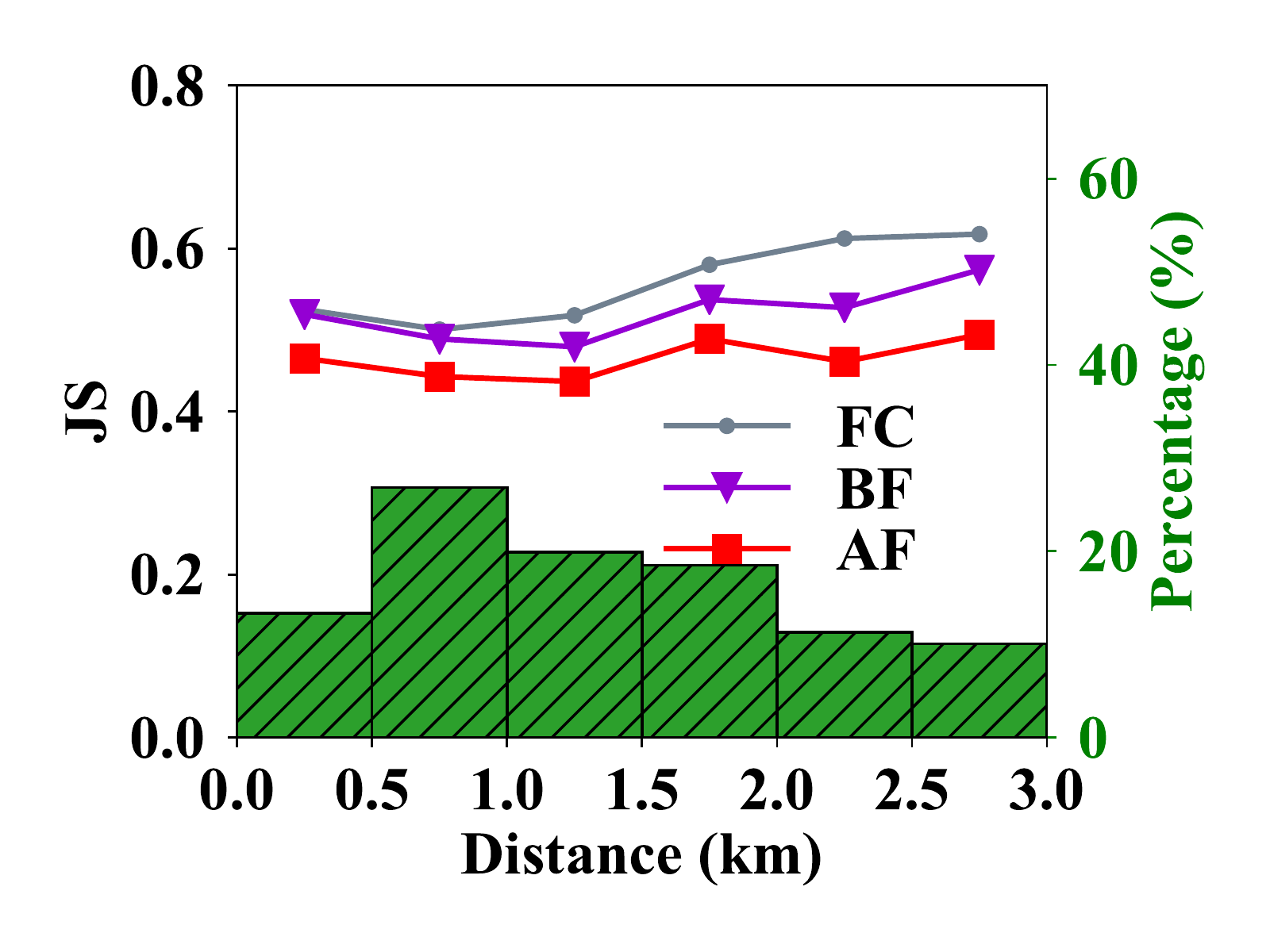}
		\end{minipage}
		\label{fig:dist_js_cd}
	}
	\caption{Effect of Distances, JS.}
	\label{fig:dist_js}
	\vspace{-10pt}
\end{figure}

In this experiment, we aim at investigating the effect of the distances between source and destination regions. We thus report the forecast accuracy with different distances. 
%
%To investigate the effect of forecasting performance over different distances between origin and destination, we show the forecast results with varying distance. 
%
Given a source and a destination region, we use the Euclidean distance between the centroids of the two regions as its corresponding distance. We group OD region pairs based on their distances into 6 groups as shown in Figures~\ref{fig:dist_emd},~\ref{fig:dist_kl}, and~\ref{fig:dist_js}. 
We only consider OD region pairs that are below 3 km because less than 1\% of the data is available for OD region pairs more than 3 km apart. 
%split distance into serveral buckets with bucket width to be 0.5~km. Since total percentage of data with distance larger than 3~km is less than 1\% and show large variance, we regard those data as outliers and removed in both datasets. 
%
%Therefore, the distance ranges that we collect data are: [0, 0.5),  [0.5, 1.0), [1.0, 1.5), [1.5, 2.5), [2.5, 3.0]. Then the forecasting results that fall into the same distance range are averaged and is shown in the results. Also, we only report FCD, FODS, GODS due to their better performance and one step ahead forecasting since their similarity.
%
Figures~\ref{fig:dist_emd},~\ref{fig:dist_kl}, and~\ref{fig:dist_js} report results on EMD, KL, and JS, respectively. %In all those figures, we continuously use dual-y axises to show both the metric results and data percentage at the same time, where the left y axis represents the metric value and the right y axis denotes the data percentage which is in green color and lies in the bottom of the figure. 

Figures~\ref{fig:dist_emd_nyc} and~\ref{fig:dist_emd_cd} show the EMD values at varying distances on NYC and CD, respectively. We observe that (1) BF and AF outperform FC for all distance settings and on both datasets; (2) AF outperforms BF by a clear margin. This again offers evidence of effectiveness of the proposed advanced framework and suggests that the best performance is achieved by contending the sparseness and by capturing spatio-temporal correlations. 
Next, when considering distances from 0.5 to 1.5, i.e., the first three points of the curves, we observe a clear descending trend in NYC, but this trend is less obvious in CD.

%
%This is becasuse the travel speed will be affected more by some traffic incidents, e.g., traffic lights. ++++ Does not make sense for CD. ++++
%+++ I did not go through the following paragraph.  +++
%
We also observe that curves start to increase 1 km on NYC as shown in Figure~\ref{fig:dist_emd_nyc}. The reason is amount of data in distance range [1.5, 3.0] decreases quickly, in turn introducing more fluctuations. We observe a subtle trend from distance range [1., 1.5) to distance range [1.5, 2.0) in Figure~\ref{fig:dist_emd_cd} for BF and AF, however, FC is much worse in distance range [1.5, 3.0]. We have similar observations for NYC regarding the KL and JS evaluation metrics, which is shown in Figures~\ref{fig:dist_kl_nyc} and~\ref{fig:dist_js_nyc}. The trend is much clearer in CD for the evaluation metrics of KL and JS as is shown in Figures~\ref{fig:dist_kl_cd} and~\ref{fig:dist_js_cd}. Therefore, another explanation of the increasing tread is that as the distance increases, the route options also increase, which makes the speed more stochastic and harder to forecast. Overall, AF achieves the best performance on both datasets regarding to EMD, KL, and JS evaluation metrics. 


%\subsubsection{Results w.r.t. Distance}
%\begin{figure}[t]
%	\centering
%	\subfigure[NYC Speed Dist KL]{
%		\begin{minipage}[b]{0.22\textwidth}
%			\includegraphics[width=1\textwidth]{./figure/Diff_methods_dist_kl.png}
%		\end{minipage}
%		\label{fig:nyc_cdf}
%	}
%	\subfigure[CD Speed Dist JS]{
%		\begin{minipage}[b]{0.22\textwidth}
%			\includegraphics[width=1\textwidth]{./figure/Diff_methods_dist_js.png}
%		\end{minipage}
%		\label{fig:cd_cdf}
%	}
%	\caption{Distance Distribution}
%	\label{fig:result_dist}
%\end{figure}
\subsubsection{Effect of Proximity Matrices}
\label{sect:exp_pm}
\begin{figure}[th]
	\centering
	\subfigure[Impact of $\sigma$]{
		\begin{minipage}[b]{0.22\textwidth}
			\includegraphics[width=1\textwidth]{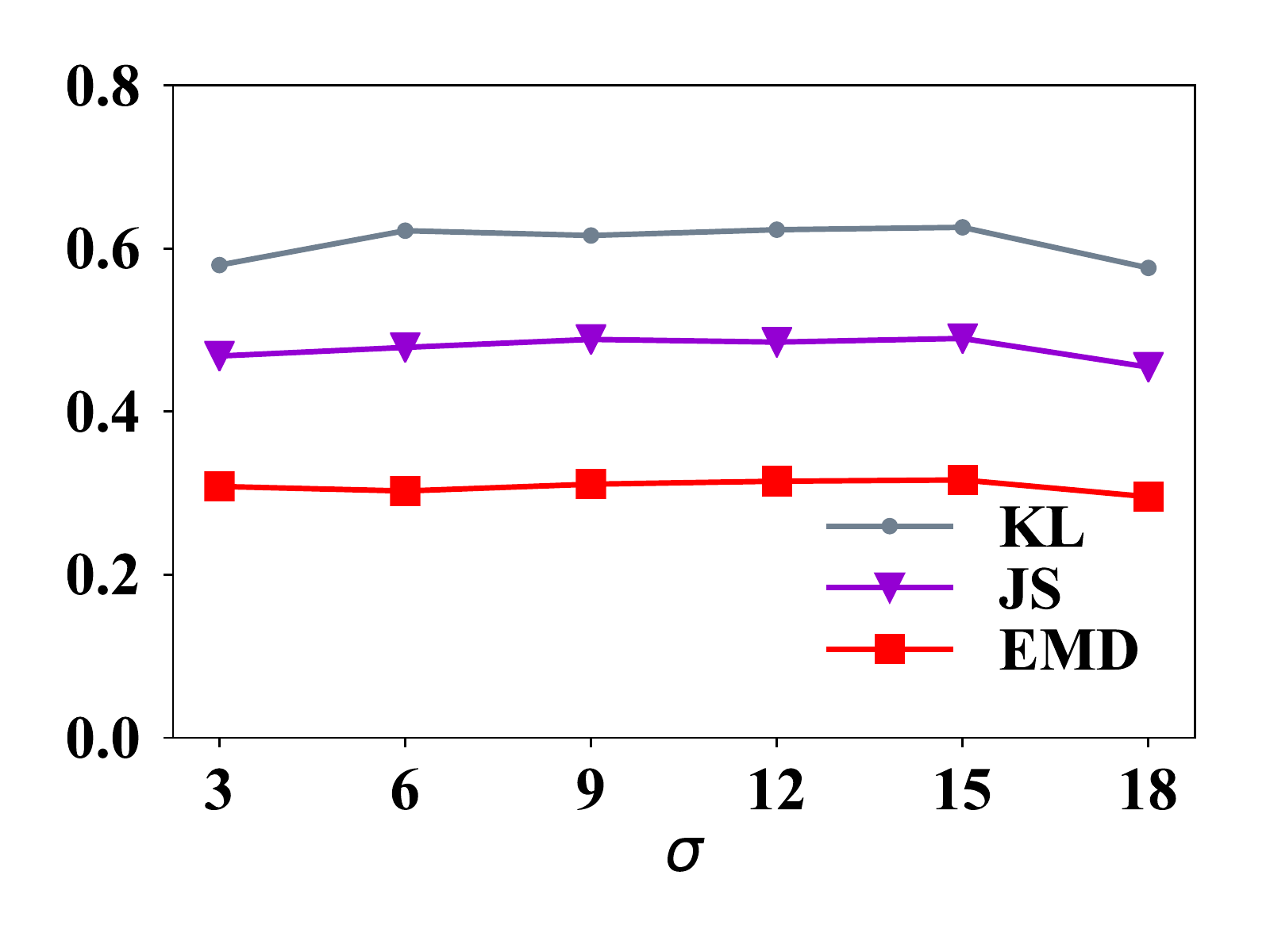}
		\end{minipage}
		\label{fig:nyc_hopk}
	}
	\subfigure[Impact of $\alpha$]{
		\begin{minipage}[b]{0.22\textwidth}
			\includegraphics[width=1\textwidth]{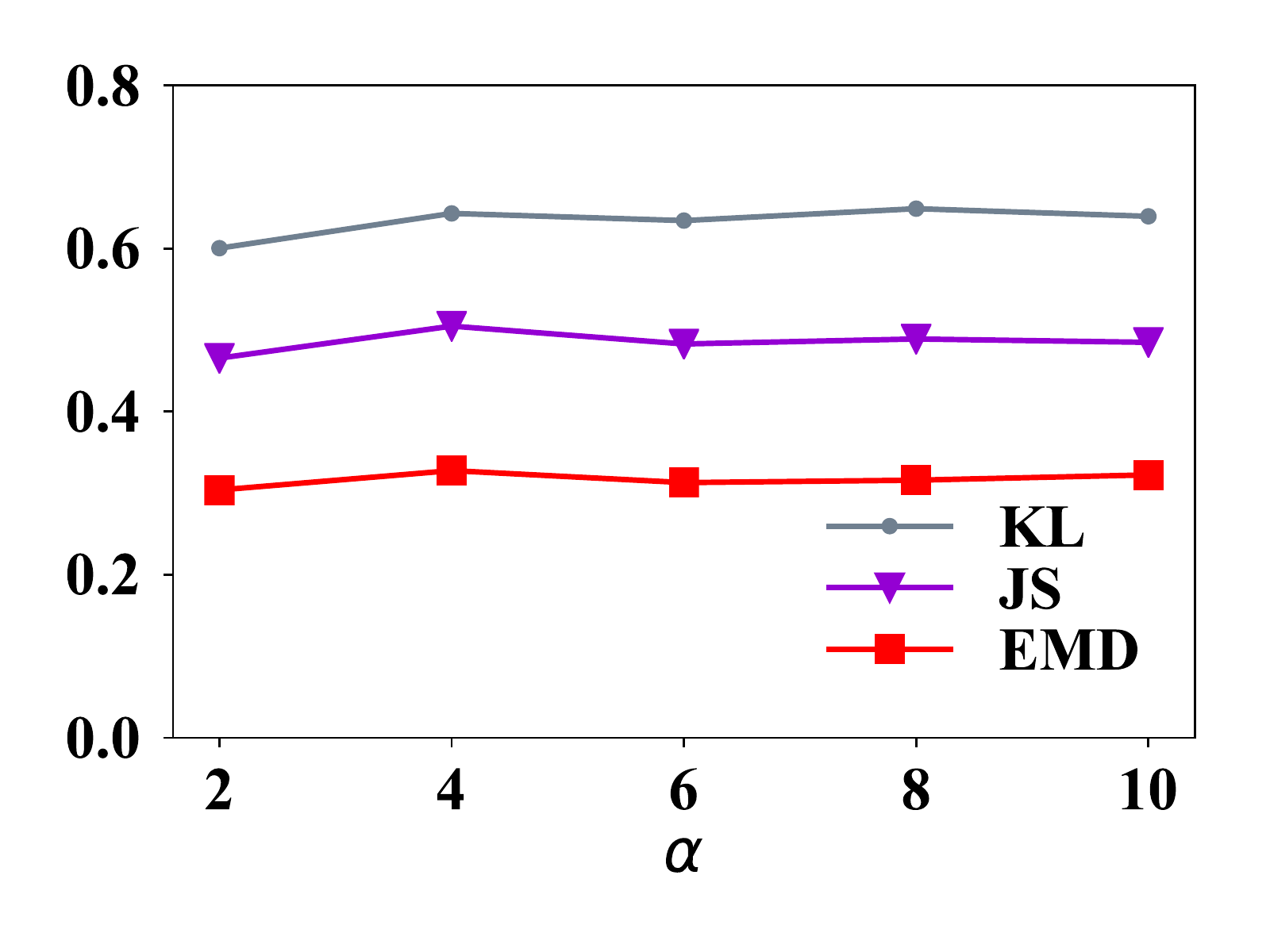}
		\end{minipage}
		\label{fig:cd_hopk}
	}
	\caption{Results w.r.t. Hop $\alpha$.}
	\label{fig:result_hopk}
	\vspace{-5pt}
\end{figure}
We conduct a last set of experiments to investigate the effect of the parameters $\sigma$ and $\alpha$ when constructing the proximity matrix $\mW$ in the advanced framework. We only report results for CD due to the space limitation and because NYC yields similar results. Figures~\ref{fig:nyc_hopk} and~\ref{fig:cd_hopk} show the accuracy when varying $\alpha$ and $\sigma$. 
The proposed AF is insensitive to $\sigma$ and $\alpha$. In other words, using proximity matrices is a robust way of capturing spatial correlations. 
%
%In all the above mentioned three evaluation metrics, we cannot observe a clear trend, i.e., the variations are very subtle and we believe that selection of $\alpha$ and $\sigma$ does not matter too much which in turn proves the robustness of capturing spatial correlations in GODS. 

\section{Conclusions and Outlook}
An increasinglly pertinent settings are asking for full OD matrices contain stochastic travel costs between any pair of regions in near future. However, instantiating such OD matrices calls for a large amount of vehicle trajectories which is almost an impossible task to fullfil in reality. We define and study the problem of stochastic origin-destination matrix forecasting in this setting. First, a data-driven, end-to-end deep learning framework is proposed to address the data sparsenss problem by taking advantage of matrix factorization and recurrent neural networks. Further, a dual-stage graph convolution is integrated into factorization and recurrent neural networks to better capture the spatial correlations and thus lift performance. Emperical studies on two real datasets from different countries, New York City and Chengdu City, demonstrate that the proposed framework outperforms other methods in all the experimental settings. 

In future work, it is of interest to extend the framework to support continuous distribution models such as Gaussian mixture models. It is also of itnerest to explore distributed and parallel computing framework~\cite{DBLP:conf/waim/YuanSWYZY10,DBLP:conf/dasfaa/YangMQZ09} to support both the pre-processing and learning when having large amount of trajectory data.  

%\begin{acks}
%  The authors would like to thank Dr. Yuhua Li for providing the
%  MATLAB code of the \textit{BEPS} method.
%
%  The authors would also like to thank the anonymous referees for
%  their valuable comments and helpful suggestions. The work is
%  supported by the \grantsponsor{GS501100001809}{National Natural
%    Science Foundation of
%    China}{http://dx.doi.org/10.13039/501100001809} under Grant
%  No.:~\grantnum{GS501100001809}{61273304}
%  and~\grantnum[http://www.nnsf.cn/youngscientists]{GS501100001809}{Young
%    Scientists' Support Program}.
%
%\end{acks}

\bibliographystyle{ACM-Reference-Format}
\bibliography{RGCN}

\end{document}